%% file: main.tex
\documentclass{article}
\pdfoutput=1
\PassOptionsToPackage{numbers, compress}{natbib}

     \usepackage[nonatbib,final]{neurips_2021}

\usepackage[table]{xcolor}
\usepackage[utf8]{inputenc} 
\usepackage[T1]{fontenc}    
\usepackage{hyperref}       
\usepackage{url}            
\usepackage{booktabs}       
\usepackage{amsfonts}       
\usepackage{nicefrac}       
\usepackage{microtype}      
\usepackage{xcolor}         
\usepackage{amsmath}
\usepackage{multirow}
\usepackage{tablefootnote}
\usepackage{pdfpages}
\usepackage[safe]{tipa}
\input{additional_libraries}

\definecolor{Gray}{gray}{0.9}
\title{Self-Supervised Learning Disentangled Group Representation as Feature}

\author{%
 \textbf{Tan Wang}\textsuperscript{1} \quad \textbf{Zhongqi Yue}\textsuperscript{1,3} \quad \textbf{Jianqiang Huang}\textsuperscript{1,3} \quad \textbf{Qianru Sun}\textsuperscript{2} \quad \textbf{Hanwang Zhang}\textsuperscript{1} \\
\small \textsuperscript{1}Nanyang Technological University\quad \textsuperscript{2}Singapore Management University \quad  \textsuperscript{3}Damo Academy, Alibaba Group\\
\tt\small \{tan317,yuez0003,hanwangzhang\}@ntu.edu.sg\\ \tt\small
  jianqiang.jqh@gmail.com \quad
qianrusun@smu.edu.sg}

\begin{document}

\maketitle

\input{0_Abstract}
\input{1_Introduction}

\input{2_Relatedwork}
\input{3_Approach}
\input{4_Justification}
\input{5_Experiments}
\input{6_Conclusion}

\bibliography{ref}

\newpage

\input{appendix/appx_main}



\end{document}

%% file: additional_libraries.tex
\usepackage{amsthm}
\theoremstyle{definition}
\newtheorem{theorem}{Theorem}
\theoremstyle{definition}

\theoremstyle{definition}

\newtheorem{lemma}{Lemma}
\theoremstyle{definition}

\newcommand{\ie}{\textit{i.e.}}
\newcommand{\eg}{\textit{e.g.}}

\newcommand{\wrt}{\textit{w.r.t. }}

\newcommand{\norm}[1]{\left\lVert#1\right\rVert}

\usepackage{enumitem}
\usepackage{mathrsfs}
\usepackage{amsmath}
\usepackage{graphicx}
\usepackage{caption}
\usepackage{wrapfig}
\usepackage{enumitem} 
\usepackage{tikz} 
\usepackage{pifont}

\usepackage{multirow}
\usepackage{booktabs}
\usepackage{bm}
\usepackage{diagbox}    

\definecolor{mygray}{gray}{0.9}

%% file: 0_Abstract.tex
\begin{abstract}
A good visual representation is an inference map from observations (images) to features (vectors) that faithfully reflects the hidden modularized generative factors (semantics). In this paper, we formulate the notion of ``good'' representation from a group-theoretic view using Higgins' definition of \emph{\textbf{disentangled representation}}~\cite{higgins2018towards}, and show that existing Self-Supervised Learning (SSL) only disentangles simple augmentation features such as rotation and colorization, thus unable to modularize the remaining semantics. To break the limitation, we propose an iterative SSL algorithm: Iterative Partition-based Invariant Risk Minimization (IP-IRM), which successfully grounds the abstract semantics and the group acting on them into concrete contrastive learning.
At each iteration, IP-IRM first partitions the training samples into two subsets that correspond to an entangled group element.
Then, it minimizes a subset-invariant contrastive loss, where the invariance guarantees to disentangle the group element. We prove that IP-IRM converges to a fully disentangled representation and show its effectiveness on various benchmarks. Codes are available at {\small \url{https://github.com/Wangt-CN/IP-IRM}}.
\end{abstract}

%% file: 1_Introduction.tex
\section{Introduction}
\input{figure1}
Deep learning is all about learning feature representations~\cite{bengio2013representation}. Compared to the conventional end-to-end supervised learning, Self-Supervised Learning (SSL) first learns a generic feature representation (\eg, a network backbone) by training with unsupervised pretext tasks such as the prevailing contrastive objective~\cite{he2019moco,chen2020simple}, and then the above stage-1 feature is expected to serve various stage-2 applications with proper fine-tuning. SSL for visual representation is so fascinating that it is the first time that we can obtain ``good'' visual features for free, just like the trending pre-training in NLP community~\cite{devlin2019bert, brown2020language}. However, most SSL works only 
care how much stage-2 performance an SSL feature can  improve, but overlook what feature SSL is learning, why it can be learned, what cannot be learned, what the gap between SSL and Supervised Learning (SL) is, and when SSL can surpass SL?

The crux of answering those questions is to formally understand \emph{what a feature representation is} and \emph{what a good one is}. We postulate the classic world model of visual generation and feature representation~\cite{anderson1972more,rao1999learning} as in Figure~\ref{fig:1}. Let $\mathcal{U}$ be a set of (unseen) \emph{semantics}, \eg, attributes such as ``digit'' and ``color''. There is a set of \emph{independent and causal mechanisms}~\cite{parascandolo2018learning} $\varphi: \mathcal{U}\to \mathcal{I}$, generating images from semantics, \eg, writing a digit ``0'' when thinking of ``0''~\cite{scholkopf2012on}. A \textbf{visual representation} is the inference process $\phi:\mathcal{I}\to \mathcal{X}$ that maps image pixels to vector space features, \eg, a neural network. We define \textbf{semantic representation} as the functional composition $f:\mathcal{U}\to \mathcal{I}\to\mathcal{X}$. In this paper, we are only interested in the parameterization of the inference process for feature extraction, but not the generation process, \ie, we assume $\forall I\in\mathcal{I}$, $\exists u\in\mathcal{U}$, such that $I = \varphi(u)$ is fixed as the observation of each image sample. Therefore, we consider semantic and visual representations the same as \textbf{feature representation}, or simply \textbf{representation}, and we slightly abuse $\phi(I) := f\left(\varphi^{-1}(I)\right)$, \ie, $\phi$ and $f$ share the same trainable parameters. We call the vector $\mathbf{x}=\phi(I)$ as \textbf{feature}, where $\mathbf{x}\in\mathcal{X}$.

We propose to use Higgins' definition of \textbf{disentangled representation}~\cite{higgins2018towards} to define what is ``good''.

\noindent\textbf{Definition 1.} (Disentangled Representation) \textit{Let $\mathcal{G}$ be the group acting on $\mathcal{U}$, \ie, $g\cdot u\in \mathcal{U}\times \mathcal{U}$ transforms $u\in \mathcal{U}$, \eg, a ``turn green'' group element changing the semantic from ``red'' to ``green''. Suppose there is a direct product decomposition\footnote[1]{Note that ${g}_i$ can also denote a cyclic subgroup $\mathcal{G}_i$ such as rotation $[0^{\circ}:1^{\circ}: 360^{\circ}]$, or a countable one but treated as cyclic such as translation $[(0,0):(1,1):(\textrm{width},\textrm{height})]$ and color $[0:1:255]$.} $\mathcal{G}={g}_1\times\ldots\times {g}_m$ and $\mathcal{U}=\mathcal{U}_1\times\ldots\times \mathcal{U}_m$, where ${g}_i$ acts on $\mathcal{U}_i$ respectively. A feature representation is disentangled if there exists a group $\mathcal{G}$ acting on $\mathcal{X}$ such that:}
\begin{enumerate}[leftmargin=+.2in]
    \item \emph{Equivariant}: \textit{$\forall g\in \mathcal{G}, \forall u\in \mathcal{U}, f(g\cdot u)=g \cdot f(u)$, \eg, the feature of the changed semantic: ``red'' to ``green'' in $\mathcal{U}$, is equivalent to directly change the color vector in $\mathcal{X}$ from ``red'' to ``green''.}
    \item \emph{Decomposable}: \textit{there is a decomposition $\mathcal{X}=\mathcal{X}_1 \times \ldots \times \mathcal{X}_m$, such that each $\mathcal{X}_i$ is fixed by the action of all ${g}_j,j\neq i$ and affected only by ${g}_i$, \eg, changing the ``color'' semantic in $\mathcal{U}$ does not affect the ``digit'' vector in $\mathcal{X}$.}
\end{enumerate}

\input{figure2}

Compared to the previous definition of feature representation which is a static mapping, the disentangled representation in Definition 1 is dynamic as it explicitly incorporate \textbf{group representation}~\cite{harris1991representation}, which is a homomorphism from group to group actions on a space, \eg, $\mathcal{G}\to \mathcal{X}\times \mathcal{X}$, and it is common to use the feature space $\mathcal{X}$ as a shorthand---this is where our title stands.

Definition 1 defines ``good'' features in the common views: 1) \emph{Robustness}: a good feature should be invariant to the change of environmental semantics, such as external interventions~\cite{ilyas2019adversarial, wang2021causal} or domain shifts~\cite{ganin2016domain}. By the above definition, a change is always retained in a subspace $\mathcal{X}_i$, while others are not affected. Hence, the subsequent classifier will focus on the invariant features and ignore the ever-changing $\mathcal{X}_i$. 2) \emph{Zero-shot Generalization}: even if a new combination of semantics is unseen in training, each semantic has been learned as features. So, the metrics of each $\mathcal{X}_i$ trained by seen samples remain valid for unseen samples~\cite{yue2021counterfactual}.

Are the existing SSL methods learning disentangled representations? No. We show in Section~\ref{sec:4} that they can only disentangle representations according to the hand-crafted augmentations, \eg, color jitter and rotation. For example, in Figure~\ref{fig:2} (a), even if we only use the augmentation-related feature, the classification accuracy of a standard SSL (SimCLR~\cite{chen2020simple}) does not lose much as compared to the full feature use. Figure 2 (b) visualizes that the CNN features in each layer are indeed entangled (\eg, tyre, motor, and background in the motorcycle image). In contrast, our approach IP-IRM, to be introduced below, disentangles more useful features beyond augmentations.

In this paper, we propose Iterative Partition-based Invariant Risk Minimization (\textbf{IP-IRM} \textipa{[\textsecstress ai\textprimstress p\textschwa:m]}) that guarantees to learn disentangled representations in an SSL fashion. We present the algorithm in Section~\ref{sec:3}, followed by the theoretical justifications in Section~\ref{sec:4}. In a nutshell, at each iteration, IP-IRM first partitions the training data into two disjoint subsets, each of which is an orbit of the already disentangled group, and the cross-orbit group corresponds to an entangled group element $g_i$. Then, we adopt the \textbf{Invariant Risk Minimization (IRM)}~\cite{arjovsky2019invariant} to implement a \textbf{partition-based} SSL, which disentangles the representation $\mathcal{X}_i$ \wrt $g_i$. Iterating the above two steps eventually converges to a fully disentangled representation \wrt $\prod_{i=1}^m {g}_i$. In Section~\ref{sec:5}, we show promising experimental results on various feature disentanglement and SSL benchmarks.

%% file: figure1.tex
\begin{wrapfigure}{r}{0.4\textwidth}
\vspace{-8mm}
\captionsetup{font=footnotesize,labelfont=footnotesize}
    \centering
    \includegraphics[width=.9\linewidth]{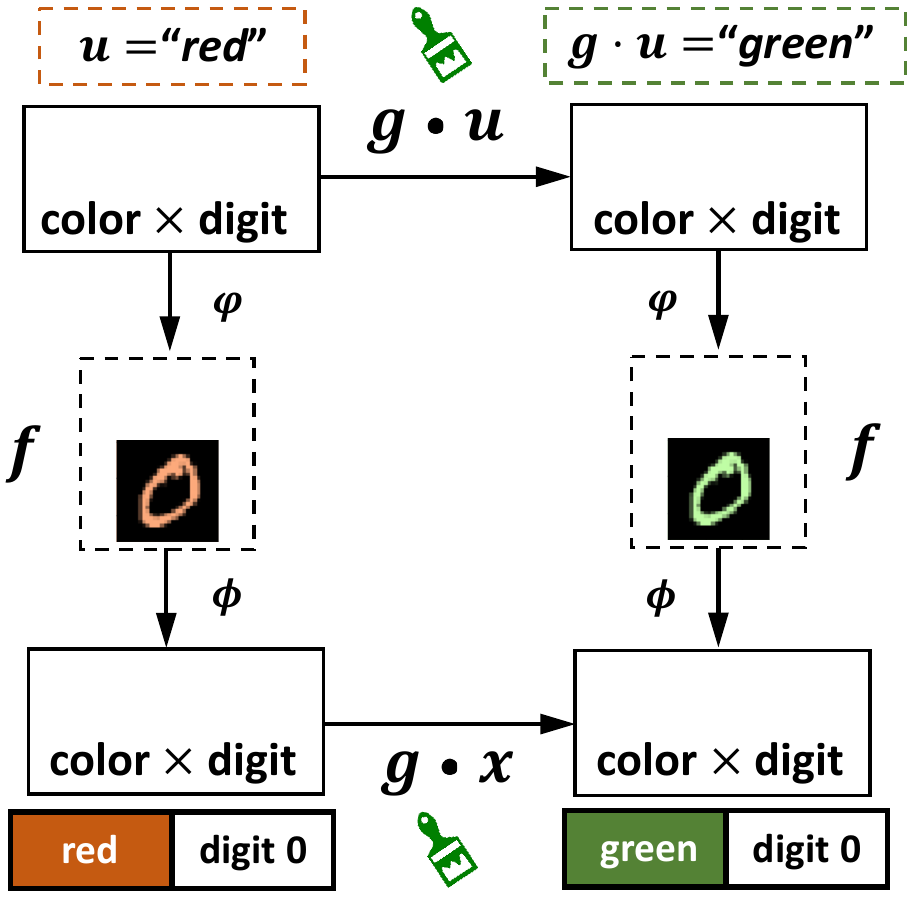}
    
    \begin{tikzpicture}[overlay, remember picture]
        \node [below left,text width=3cm,align=center] at (0.05,4.77) {\scalebox{1.2}{$\boldsymbol{\mathcal{U}}$}};
        \node [below left,text width=3cm,align=center] at (3.05,4.77) {\scalebox{1.2}{$\boldsymbol{\mathcal{U}}$}};
        
        \node [below left,text width=3cm,align=center] at (0.05,3.37) {\scalebox{1.2}{$\boldsymbol{\mathcal{I}}$}};
        \node [below left,text width=3cm,align=center] at (3.05,3.37) {\scalebox{1.2}{$\boldsymbol{\mathcal{I}}$}};
        
        \node [below left,text width=3cm,align=center] at (0.05,1.77) {\scalebox{1.2}{$\boldsymbol{\mathcal{X}}$}};
        \node [below left,text width=3cm,align=center] at (3.05,1.77) {\scalebox{1.2}{$\boldsymbol{\mathcal{X}}$}};
    \end{tikzpicture}
    
    \caption{Disentangled representation is an equivariant map between the semantic space $\mathcal{U}$ and the vector space $\mathcal{X}$, which is decomposed into ``color'' and ``digit''.}
    \label{fig:1}
    \vspace{-4mm}
\end{wrapfigure}

%% file: figure2.tex
\begin{figure}
    \centering
    \captionsetup{font=footnotesize,labelfont=footnotesize}
    \includegraphics[width=.98\linewidth]{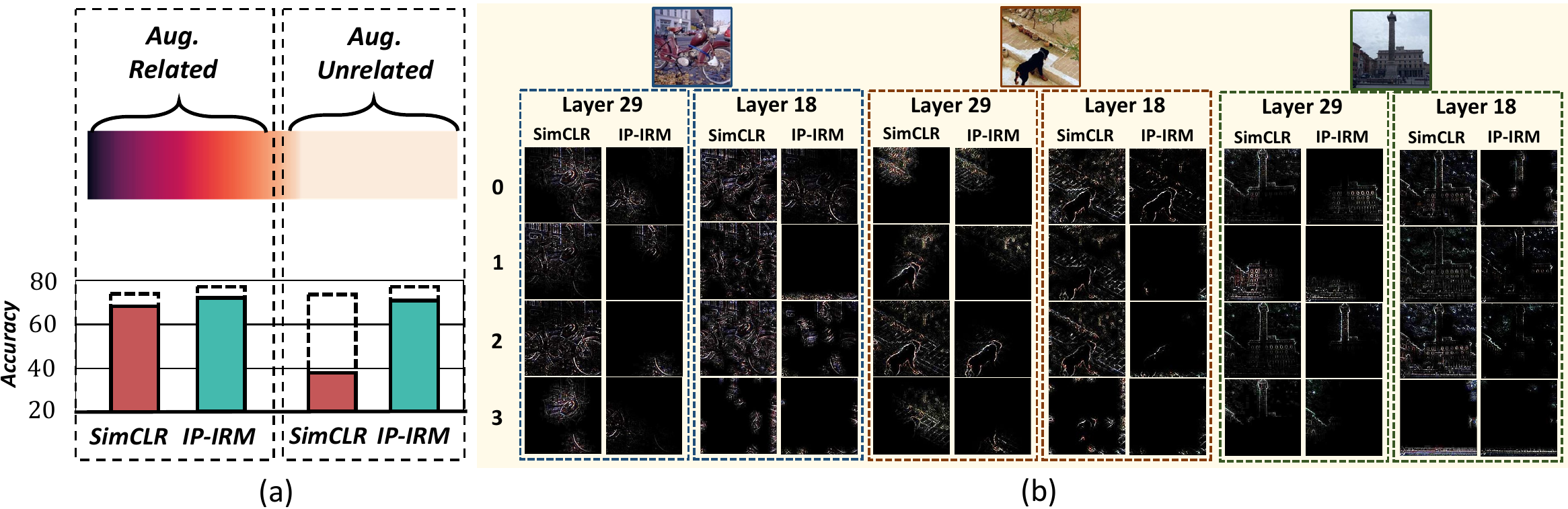}
    \caption{(a) The heat map visualizes feature dimensions related to augmentations (aug. related) and unrelated to augmentations (aug. unrelated), whose respective classification accuracy is shown in the bar chart below. Dashed bar denotes the accuracy using full feature dimensions. Experiment was performed on STL10~\cite{coates2011analysis} with representation learnt with SimCLR~\cite{chen2020simple} and our IP-IRM. (b) Visualization of CNN activations~\cite{springenberg2014striving} of 4 filters on layer 29 and 18 of VGG~\cite{karen2015very} trained on ImageNet100~\cite{tian2020contrastive}. The filters were chosen by first clustering the aug. unrelated filters with $k$-means ($k=4$) and then selecting the filters corresponding to the cluster centers.}
    \label{fig:2}
    \vspace{-4mm}
\end{figure}

%% file: 2_Relatedwork.tex
\section{Related Work}

\noindent\textbf{Self-Supervised Learning}. SSL aims to learn representations from 
unlabeled data
with hand-crafted pretext tasks~\cite{doersch2015unsupervised,noroozi2016unsupervised,gidaris2018unsupervised}.
Recently, Contrastive learning~\cite{oord2018representation, misra2020self, henaff2020data, tian2019contrastive, chen2020simple} prevails in most state-of-the-art methods. The key is to map positive samples closer, while pushing apart negative ones in the feature space. Specifically, the positive samples are from the augmented views~\cite{tian2020makes, bachman2019learning, ye2019unsupervised, hjelm2018learning} of each instance and the negative ones are other instances.
Along this direction, follow-up methods are mainly four-fold: 1) Memory-bank~\cite{wu2018unsupervised, misra2020self, he2019moco, chen2020mocov2}: storing the prototypes of all the instances computed previously into a memory bank to benefit from a large number of negative samples. 2) Using siamese network~\cite{bromley1993signature} to avoid representation collapse~\cite{grill2020bootstrap, chen2020exploring, tian2021understanding}. 3) Assigning clusters to samples to integrate inter-instance similarity into contrastive learning~\cite{caron2018deep, caron2019unsupervised, caron2020unsupervised, wang2020unsupervised, li2020prototypical}. 4) Seeking hard negative samples with adversarial training or better sampling strategies~\cite{robinson2020contrastive, chuang2020debiased, hu2020adco, kalantidis2020hard}. In contrast, our proposed IP-IRM jumps out of the above frame and introduces the \textit{disentangled representation} into SSL with group theory to show the limitations of existing SSL and how to break through them.

\noindent\textbf{Disentangled Representation}. This notion dates back to~\cite{bengio2009learning}, and henceforward becomes a high-level goal of separating the factors of variations in the data~\cite{Tran_2017_CVPR,suter2019robustly,van2019disentangled,Locatello2020disentangling}. Several works aim to provide a more precise description~\cite{do2019theory,eastwood2018framework,ridgeway2018learning} by adopting an information-theoretic view~\cite{chen2016infogan,do2019theory} and measuring the properties of a disentangled representation explicitly~\cite{eastwood2018framework,ridgeway2018learning}. We adopt the recent group-theoretic definition from Higgins \textit{et al.}~\cite{higgins2018towards}, which not only unifies the existing, but also resolves the previous controversial points~\cite{suter2018interventional,locatello2019challenging}.
Although supervised learning of disentangled representation is a well-studied field~\cite{zhu2014multi,hsieh2018learning,cai2019learning,reed2014learning,karaletsos2015bayesian}, unsupervised disentanglement based on GAN~\cite{chen2016infogan,ojha2020elastic,lin2020infogan,ren2021generative} or VAE~\cite{Higgins2017betaVAELB,chen2018isolating,zhu2020s3vae,kim2018disentangling} is still believed to be theoretically challenging~\cite{locatello2019challenging}. Thanks to the Higgins' definition, we prove that the proposed IP-IRM converges with full-semantic disentanglement using group representation theory. Notably, IP-IRM learns a disentangled representation with an inference process, without using generative models as in all the existing unsupervised methods, making IP-IRM applicable even on large-scale datasets.

\noindent\textbf{Group Representation Learning}. A group representation has two elements~\cite{Judson1994Abstract,harris1991representation}: 1) a homomorphism (\eg, a mapping function) from the group to its group action acting on a vector space, and 2) the vector space. Usually, when there is no ambiguity, we can use either element as the definition. Most existing works focus on learning the first element. They first define the group of interest, such as spherical rotations~\cite{cohen2018spherical} or image scaling~\cite{worrall2019deep,sosnovik2019scale}, and then learn the parameters of the group actions~\cite{cohen2014learning,jaegle2016understanding,quessard2020learning}. In contrast, we focus on the second element; more specifically, we are interested in learning a map between two vector spaces: image pixel space and feature vector space. Our representation learning is flexible because it delays the group action learning to downstream tasks on demand. For example, in a classification task, a classifier can be seen as a group action that is invariant to class-agnostic groups but equivariant to class-specific groups (see Section~\ref{sec:4}).

%% file: 3_Approach.tex
\section{IP-IRM Algorithm}
\label{sec:3}

\noindent\textbf{Notations}. Our goal is to learn the feature extractor $\phi$ in a self-supervised fashion. We define a partition matrix $\mathbf{P}\in \{0,1\}^{N\times 2}$ that partitions $N$ training images into $2$ disjoint subsets. $P_{i,k}=1$ if the $i$-th image belongs to the $k$-th subset and $0$ otherwise. Suppose we have a pretext task loss function $\mathcal{L}(\phi,\theta=1,k,\mathbf{P})$
defined on the samples in the $k$-th subset, where $\theta=1$ is a ``dummy'' parameter used to evaluate the invariance of the SSL loss across the subsets (later discussed in Step 1).
For example, $\mathcal{L}$ can be defined as:
\begin{equation}
    \mathcal{L}(\phi,\theta=1,k,\mathbf{P}) = \sum\limits_{\mathbf{x}\in\mathcal{X}_k}-\mathrm{log} \frac{\mathrm{exp}\left(\mathbf{x}^T \mathbf{x}^* \cdot \theta\right)}{\sum_{\mathbf{x}' \in \mathcal{X}_k \cup\mathcal{X}^*\setminus \mathbf{x}} \mathrm{exp}\left( \mathbf{x}^T \mathbf{x}' \cdot \theta\right)},
    \label{eq:contrastive_loss}
\end{equation}
where $\mathcal{X}_k = \phi(\{I_i | P_{i,k}=1\})$, and $\mathbf{x}^*\in\mathcal{X}^*$ is the augmented view feature of $\mathbf{x}\in\mathcal{X}_k$.

\noindent\textbf{Input}. $N$ training images. Randomly initialized $\phi$. A partition matrix $\mathbf{P}$ initialized such that the first column of $\mathbf{P}$ is 1, \ie, all samples belong to the first subset. Set $\mathcal{P}=\{\mathbf{P}\}$.

\noindent\textbf{Output}. Disentangled feature extractor $\phi$.

\noindent\textbf{Step 1 [Update $\phi$]}. 
 We update $\phi$ by:
\begin{equation}
    \mathop{\mathrm{min}}_{\phi} \sum_{\mathbf{P}\in \mathcal{P}} \sum_{k=1}^2 \left[ \mathcal{L}(\phi,\theta=1, k,\mathbf{P}) + \lambda_1 \norm{\nabla_{\theta=1} \mathcal{L} (\phi,\theta=1,k,\mathbf{P})}^2 \right],
    \label{eq:min}
\end{equation}
where $\lambda_1$ is a hyper-parameter. The second term delineates how far the contrast in one subset is from a constant baseline $\theta = 1$. The minimization of both of them encourages $\phi$ in different subsets close to the same baseline, \ie, invariance across the subsets. See IRM~\cite{arjovsky2019invariant} for more details. In particular, the first iteration corresponds to the standard SSL with $\mathcal{X}_1$ in Eq.~\eqref{eq:contrastive_loss} containing all training images.

\noindent\textbf{Step 2 [Update $\mathbf{P}$]}. We fix $\phi$ and find a new partition $\mathbf{P}^*$ by
\begin{equation}
    \mathbf{P}^* = \mathop{\mathrm{arg\,max}}_{\mathbf{P}} \sum_{k=1}^2 \left[ \mathcal{L} (\phi,\theta=1,k,\mathbf{P}) + \lambda_2 \norm{\nabla_{\theta=1} \mathcal{L} (\phi,\theta=1,k,\mathbf{P}) }^2 \right],
    \label{eq:max}
\end{equation}
where $\lambda_2$ is a hyper-parameter. In practice, we use a continuous partition matrix in $\mathbb{R}^{N\times 2}$ during optimization and then threshold it to $\{0,1\}^{N\times 2}$.

We update $\mathcal{P}\leftarrow \mathcal{P}\cup \mathbf{P}^*$ and iterate the above two steps until convergence.

%% file: 4_Justification.tex
\section{Justification}
\label{sec:4}

Recall that IP-IRM uses training sample \textbf{partitions} to learn the disentangled representations \wrt $\prod^m_{i=1} {g}_i$. As we have a $\mathcal{G}$-equivariant feature map between the sample space $\mathcal{I}$ and feature space $\mathcal{X}$ (the equivariance is later guaranteed by Lemma~\ref{lem:1}), we slightly abuse the notation by using $\mathcal{X}$ to denote both spaces. Also, we assume that $\mathcal{X}$ is a \textbf{homogeneous} space of $\mathcal{G}$, \ie, any sample $\mathbf{x}' \in \mathcal{X}$ can be transited from another sample $\mathbf{x}$ by a group action $g\cdot \mathbf{x}$. Intuitively, $\mathcal{G}$ is all you need to describe the diversity of the training set. It is worth noting that $g$ is any group element in $\mathcal{G}$ while ${g}_i$ is a Cartesian ``building block'' of $\mathcal{G}$, \eg, $g$ can be decomposed by $(g_1, g_2, ..., g_m)$.

We show that partition and group are tightly connected by the concept of \textbf{orbit}. Given a sample $\mathbf{x}\in\mathcal{X}$, its group orbit \wrt $\mathcal{G}$ is a sample set $\mathcal{G}(\mathbf{x}) = \{g\cdot \mathbf{x} \mid g \in \mathcal{G}\}$. As shown in Figure~\ref{fig:3} (a), if $\mathcal{G}$ is a set of attributes shared by classes, \eg, ``color'' and ``pose '', the orbit is the sample set of 
the class of $\mathbf{x}$; in Figure~\ref{fig:3} (b), if $\mathcal{G}$ denotes augmentations, the orbit is the set of augmented images. In particular, we can see that the disjoint orbits in Figure~\ref{fig:3} naturally form a partition. Formally, we have the following definition:

\noindent\textbf{Definition 2.} (Orbit \& Partition~\cite{Judson1994Abstract}) \textit{Given a subgroup $\mathcal{D}\subset\mathcal{G}$, it partitions  $\mathcal{X}$ into the disjoint subsets: $\{\mathcal{D}(c_1\cdot \mathbf{x}), ..., \mathcal{D}(c_k\cdot\mathbf{x})\}$,  where $k$ is the number of cosets $\{c_1\mathcal{D}, ..., c_k\mathcal{D}\}$, and the cosets form a factor group\footnote{Given $\mathcal{G}=\mathcal{D}\times \mathcal{K}$ with $\mathcal{K}=c_1\times \ldots \times c_k$, then $\bar{\mathcal{D}}=\{(d,e) \mid d\in\mathcal{D}\}$ is a normal subgroup of $\mathcal{G}$, and $\mathcal{G} / \bar{\mathcal{D}}$ is isomorphic to $\mathcal{K}$~\cite{Judson1994Abstract}. We write $\mathcal{G} / \mathcal{D} = \{c_i\}^k_{i=1}$ with slight abuse of notation.} $\mathcal{G} / \mathcal{D} = \{c_i\}^k_{i=1}$. In particular, $c_i\cdot\mathbf{x}$ can be considered as a sample of the $i$-th class, transited from any sample $\mathbf{x}\in\mathcal{X}$}.

Interestingly, the partition offers a new perspective for the training data format in Supervised Learning (SL) and Self-Supervised Learning (SSL). In SL, as shown in Figure~\ref{fig:3} (a), the data is labeled with $k$ classes, each of which is an orbit with $\mathcal{D}(c_i\cdot\mathbf{x})$ training samples, whose variations are depicted by the class-sharing attribute group $\mathcal{D}$. The cross-orbit group action, \eg, $c_{\textrm{dog}}\cdot\mathbf{x}$, can be read as ``turn $\mathbf{x}$ into a dog'' and such ``turn'' is always valid due to the assumption that $\mathcal{X}$ is a homogeneous space of $\mathcal{G}$. 
In SSL, as shown in Figure~\ref{fig:3} (b), each training sample $\mathbf{x}$ is augmented by the group $\mathcal{D}$. So, $\mathcal{D}(c_i\cdot\mathbf{x})$ consists of all the augmentations of the $i$-th sample, where the cross-orbit group action $c_i\cdot \mathbf{x}$ can be read as ``turn $\mathbf{x}$ into the $i$-th sample''.

Thanks to the orbit and partition view of training data, we are ready to revisit model \textbf{generalization} in a group-theoretic view by using \textbf{invariance} and \textbf{equivariance}---the two sides of the coin, whose name is \textbf{disentanglement}. For SL, we expect that a good feature is disentangled into a class-agnostic part and a class-specific part: the former (latter) is invariant (equivariant) to $\mathcal{G} / \mathcal{D}$---cross-orbit traverse, but equivariant (invariant) to $\mathcal{D}$---in-orbit traverse. By using such feature, a model can generalize to diverse testing samples (limited to $|\mathcal{D}|$ variations) by only keeping the class-specific feature. Formally, we prove that we can achieve such disentanglement by contrastive learning:

\input{figure3}

\begin{lemma} (Disentanglement by Contrastive Learning)
\textit{Training loss $-\log \frac{\exp (\mathbf{x}^T_i\mathbf{x}_j)}{\sum\nolimits_{\mathbf{x}\in\mathcal{X}}\exp(\mathbf{x}^T_j\mathbf{x})}$ disentangles $\mathcal{X}$ \wrt $(\mathcal{G} / \mathcal{D})\times\mathcal{D}$, where $\mathbf{x}_i$ and $\mathbf{x}_j$ are from the same orbit}.
\label{lem:1}
\end{lemma}

We can draw the following interesting corollaries from Lemma~\ref{lem:1} (details in Appendix):
\begin{enumerate}[leftmargin=+.2in]

\item If we use all the samples in the denominator of the loss, we can approximate to $\mathcal{G}$-equivariant features given limited training samples. This is because the loss minimization guarantees $\forall (\mathbf{x}_i, \mathbf{x}_j)\in\mathcal{X}\times\mathcal{X}, i\neq j\to \mathbf{x}_i\neq\mathbf{x}_j$, \ie, any pair corresponds to a group action.

\item Conventional cross-entropy loss in SL is a special case, if we define $\mathbf{x}\in\mathcal{X} = \{\mathbf{x}_1, ..., \mathbf{x}_k\}$ as $k$ classifier weights. So, SL does not guarantee the disentanglement of $\mathcal{G} / \mathcal{D}$, which causes generalization error if the class domain of downstream task is different from SL pre-training, \eg, a subset of $\mathcal{G} / \mathcal{D}$. 

\item In contrastive learning based SSL, $\mathcal{D} = \textrm{``augmentations''}$ (recall Figure~\ref{fig:2}), and the number of augmentations $|\mathcal{D}_{\textrm{aug}}|$ is generally much smaller compared to the class-wise sample diversity $|\mathcal{D}_{\textrm{SL}}|$ in SL. This enables the SL model to generalize to more diverse testing samples ($|\mathcal{D}_{\textrm{SL}}|$) by filtering out the class-agnostic features (\eg, background) and focusing on the class-specific ones (\eg, foreground), which explains why SSL is worse than SL in downstream classification.

\item In SL, if the number of training samples per orbit is not enough, \ie, smaller than $|\mathcal{D}(c_i\cdot \mathbf{x})|$, the disentanglement between $\mathcal{D}$ and $\mathcal{G} / \mathcal{D}$ cannot be guaranteed, such as the challenges in few-shot learning~\cite{yue2020interventional}. Fortunately, in SSL, the number is enough as we always include all the augmented samples in training. Moreover, we conjecture that $\mathcal{D}_{\textrm{aug}}$ only contains simple cyclic group elements such as rotation and colorization, which are easier for representation learning.
\end{enumerate}

Lemma~\ref{lem:1} does not guarantee the decomposability of each $d\in \mathcal{D}$.
Nonetheless, the downstream model can still generalize by keeping the class-specific features affected by $\mathcal{G} / \mathcal{D}$.
Therefore, the key to fill the gap or even let SSL surpass SL is to achieve the full disentanglement of $\mathcal{G} / \mathcal{D}_{\textrm{aug}}$.

\begin{theorem}
\textit{The representation is fully disentangled \wrt $\mathcal{G} / \mathcal{D}_{\textnormal{aug}}$ if and only if $\forall c_i \in \mathcal{G} / \mathcal{D}_{\textnormal{aug}}$, the contrastive loss in Eq.~\eqref{eq:contrastive_loss} is invariant to the 2 orbits of partition $\{\mathcal{G}'(c_i\cdot \mathbf{x}), \mathcal{G}'(c_i^{-1}\cdot \mathbf{x})\}$, where $\mathcal{G}'= \mathcal{G} / c_i = \mathcal{D}_{\textrm{\textnormal{aug}}} \times c_1 \times \ldots \times c_{i-1} \times c_{i+1} \times \ldots \times c_k$}.
\label{thm:1}
\end{theorem}

The maximization in \textbf{Step 2} is based on the contra-position of the sufficient condition of Theorem~\ref{thm:1}. Denote the currently disentangled group as $\mathcal{D}$ (initially $\mathcal{D}_{\textrm{aug}}$).
If we can find a partition $\mathbf{P}^*$ to maximize the loss in Eq.~\eqref{eq:max}, \ie, SSL loss is variant across the orbits, then $\exists h \in \mathcal{G} / \mathcal{D}$ such that the representation of $h$ is entangled, \ie, $\mathbf{P}^*=\{ \mathcal{D}(h \cdot \mathbf{x}), \mathcal{D}(h^{-1} \cdot \mathbf{x}) \}$. Figure~\ref{fig:3} (c) illustrates a discovered partition about color.
The minimization in \textbf{Step 1} is based on the necessary condition of Theorem~\ref{thm:1}. Based on the discovered $\mathbf{P}^*$, if we minimize Eq.~\eqref{eq:min}, we can further disentangle $h$ and update $\mathcal{D}\leftarrow \mathcal{D} \times h$.
Overall, IP-IRM converges as $\mathcal{G} / \mathcal{D}_{\textrm{aug}}$ is finite. Note that an improved contrastive objective~\cite{xiao2020should} can further disentangle each $d\in \mathcal{D}_{\textrm{aug}}$ and achieve full disentanglement \wrt $\mathcal{G}$.

%% file: figure3.tex
\begin{figure}
    \centering
    \captionsetup{font=footnotesize,labelfont=footnotesize}
    \includegraphics[width=.98\linewidth]{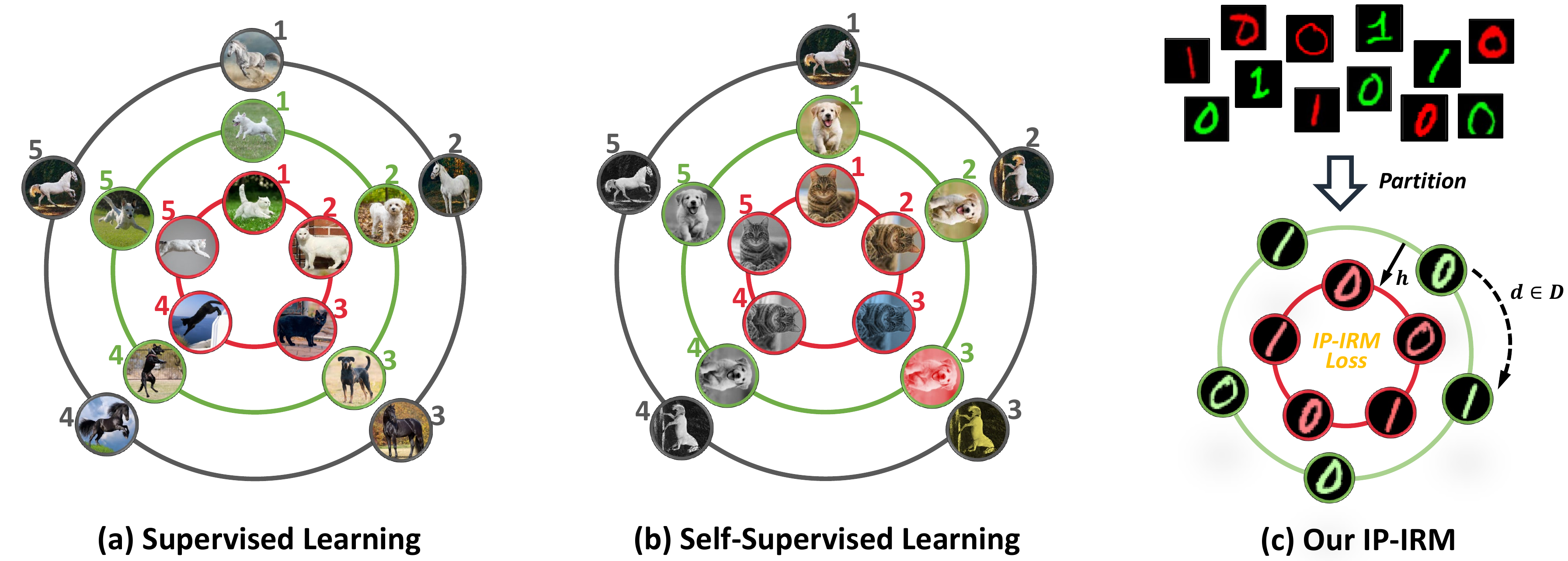}
    \caption{Each orbit only illustrates with 5 samples. (a) Orbit: the training samples of a class; $\mathcal{D}$ in-orbit actions: intra-class variations ($1\!\!\rightarrow\!\! 2$:standing, $2\!\!\rightarrow\!\!3$:blacken, $3\!\!\rightarrow\!\!4$:jumping, $4\!\!\rightarrow\!\!5$:whiten, $5\!\!\rightarrow\!\!1$:running); $\mathcal{G} / \mathcal{D}$ cross-orbit actions: inter-class variations. (b) Orbit: a sample and its augmented samples; $\mathcal{D}$ in-orbit actions: augmentations ($1\!\!\rightarrow\!\! 2$:clock-wise rotation, $2\!\!\rightarrow\!\!3$:color jitter, $3\!\!\rightarrow\!\!4$:gray scale, $4\!\!\rightarrow\!\!5$:counterclockwise rotation, $5\!\!\rightarrow\!\!1$:color); $\mathcal{G} / \mathcal{D}$ cross-orbit actions: inter-sample variations.  (c) Step 2 in IP-IRM discovers 2 orbits, where the cross-orbit action corresponds to a group action ``green to red'' or ``red to green'', which is yet disentangled.}
    \label{fig:3}
    \vspace{-3mm}
\end{figure}

%% file: 5_Experiments.tex
\section{Experiments}

\label{sec:5}

\subsection{Unsupervised Disentanglement}

\noindent\textbf{Datasets}. We used two datasets. \textbf{CMNIST}~\cite{arjovsky2019invariant} has 60,000 digit images with semantic labels of digits (0-9) and colors (red and green). These images differ in other semantics (\eg, slant and font) that are not labeled. Moreover, there is a strong correlation between digits and colors (most 0-4 in red and 5-9 in green), increasing the difficulty to disentangle them.
\textbf{Shapes3D}~\cite{kim2018disentangling} contains 480,000 images with 6 labelled semantics, \ie, size, type, azimuth, as well as floor, wall and object color. Note that we only considered the \emph{first three} semantics for evaluation, as the standard augmentations in SSL will contaminate any color-related semantics.

\noindent\textbf{Settings}. We adopted 6 representative disentanglement metrics: \textit{Disentangle Metric for Informativeness} (DCI)~\cite{eastwood2018framework}, \textit{Interventional Robustness Score} (IRS)~\cite{suter2019robustly}, \textit{Explicitness Score} (EXP)~\cite{ridgeway2018learning}, \textit{Modularity Score} (MOD)~\cite{ridgeway2018learning} and the accuracy of predicting the ground-truth semantic labels by two classification models called
\emph{logistic regression} (LR) and \emph{gradient boosted trees} (GBT)~\cite{locatello2019challenging}.
Specifically, DCI and EXP measure the explicitness, \ie, the values of semantics can be decoded from the feature using a linear transformation. 
MOD and IRS measure the modularity, \ie, 
whether each feature dimension is equivariant to the shift of a single semantic. See Appendix for more detailed formula of the metrics.
In evaluation, we trained CNN-based feature extractor backbones with comparable number of parameters for all the baselines and our IP-IRM.
The full implementation details are in Appendix.

\input{disentangle}

\noindent\textbf{Results}.
In Table~\ref{tab:cmnist}, 
we compared the proposed IP-IRM to the standard SSL method SimCLR~\cite{chen2020simple} as well as several generative disentanglement methods~\cite{kingma2014ICLR,higgins2016beta,burgess2018understanding,chen2018isolating,kim2018disentangling}.
On both CMNIST and Shapes3D dataset, IP-IRM outperforms SimCLR regarding all metrics except for only IRS where the most relative gain is 8.8\% for MOD.
For this MOD, we notice that VAE performs better than our IP-IRM by 6 points, \ie, 0.82 v.s. 0.76 for Shapes3D. 
This is because VAE explicitly pursues a high modularity score through regularizing the dimension-wise independence in the feature space. 
However, this regularization is adversarial to discriminative objectives~\cite{chen2018zero,yue2021counterfactual}.
Indeed, we can observe from the column of LR (\ie, the performance of downstream linear classification) that VAE methods have clearly poor performance especially on the more challenging dataset Shapes3D.
We can draw the same conclusion from the results of GBT.
Different from VAE methods, our IP-IRM is optimized towards disentanglement without such regularization, and is thus able to outperform the others in downstream tasks while obtaining a competitive value of modularity.

\noindent\textbf{What do IP-IRM features look like?}
Figure~\ref{fig:tsne} visualizes the features learned by SimCLR and our IP-IRM on two datasets: CMNIST in Figure~\ref{fig:tsne} (a) and STL10 dataset in Figure~\ref{fig:tsne} (b). 
In the following, we use Figure~\ref{fig:tsne} (a) as the example, and can easily draw the similar conclusions from Figure~\ref{fig:tsne} (b).
On the left-hand side of Figure~\ref{fig:tsne} (a), it is obvious that there is no clear boundary to distinguish the semantic of color in the SimCLR feature space. 
Besides, the features of the same digit semantic are scattered in two regions. 
%
On the right-hand side of (a), we have 3 observations for IP-IRM. 
1) The features are well clustered and each cluster corresponds to a specific semantic of either digit or color.
This validates the \emph{equivariant} property of IP-IRM representation that it responds to any changes of the existing semantics, \eg, digit and color on this dataset.
%
2) The feature space has the symmetrical structure for each individual semantic, validating the \emph{decomposable} property of IP-IRM representation. More specifically, i) mirroring a feature (\wrt ``*'' in the figure center) indicates the change on the only semantic of color, regardless of the other semantic (digit); and ii) a counterclockwise rotation (denoted by black arrows from same-colored 1 to 7) indicates the change on the only semantic of digit.
3) IP-IRM reveals the true distribution (similarity) of different classes. For example, digits 3, 5, 8 sharing sub-parts (curved bottoms and turnings) have closer feature points in the IP-IRM feature space.

\input{tsne}

\input{dim_group}

\noindent\textbf{How does IP-IRM disentangle features?}
\textit{1) Discovered $\mathbf{P}^*$}: To visualize the discovered partitions $\mathbf{P}^*$ at each maximization step, we performed an experiment on a binary CMNIST (digit 0 and 1 in color red and green), and show the results in Figure~\ref{fig:dim_group} (a).
Please kindly refer to Appendix for the full results on CMNIST.
First, each partition tells apart a specific semantic into two subsets, \eg, in Partition \#1, red and green digits are separated. 
Second, besides the obvious semantics---digit and color (labelled on the dataset), we can discover new semantics, \eg, the digit slant shown in Partition \#3.
\textit{2) Disentangled Representation}:
In Figure~\ref{fig:dim_group} (b), we aim to visualize how equivariant each feature dimension is to the change of each semantic, \ie, a darker color shows that a dimension is more equivariant \wrt the semantic indicated on the left.
We can see that SimCLR fails to learn the decomposable representation, \eg, the 8-th dimension captures azimuth, type and size in Shapes3D.
In contrast, our IP-IRM achieves disentanglement by representing the semantics into interpretable dimensions,
\eg, the 6-th and 7-th dimensions captures the size, the 4-th for type and the 2-nd and 9-th for azimuth on the Shapes3D.
Overall, the results support the justification in Section~\ref{sec:4}, \ie, we discover a new semantic (affected by $h$) through the partition $\mathbf{P}^*$ at each iteration and IP-IRM eventually converges with a disentangled representation.

\input{5_Experiments_part2}

%% file: disentangle.tex
\begin{table}[t!]
\centering
\captionsetup{font=footnotesize,labelfont=footnotesize,skip=5pt}
\setlength\extrarowheight{1pt}
\scalebox{0.75}{
\begin{tabular}{p{0.5cm}<{\centering} p{2.7cm} p{1.6cm}<{\centering}p{1.6cm}<{\centering}p{1.6cm}<{\centering}p{1.6cm}<{\centering}p{1.6cm}<{\centering}p{1.6cm}<{\centering}p{1.6cm}<{\centering}}
\hline\hline
& Method & DCI & IRS & MOD & EXP & LR & GBT & Average\\
\hline

\multicolumn{1}{c}{\multirow{7}{*}{\rotatebox{90}{\textbf{CMNIST}}}} & VAE~\cite{kingma2014ICLR} & \textbf{0.948}\scalebox{.8}{$\pm$0.004} & - & 0.664\scalebox{.8}{$\pm$0.121} & 0.968\scalebox{.8}{$\pm$0.007} & 0.824\scalebox{.8}{$\pm$0.019} & \textbf{0.948}\scalebox{.8}{$\pm$0.004} & 0.849\scalebox{.8}{$\pm$0.057}\\
& $\beta$-VAE~\cite{higgins2016beta} & 0.945\scalebox{.8}{$\pm$0.002} & - & 0.705\scalebox{.8}{$\pm$0.073} & 0.963\scalebox{.8}{$\pm$0.006} & 0.809\scalebox{.8}{$\pm$0.013} & 0.945\scalebox{.8}{$\pm$0.003} & 0.874\scalebox{.8}{$\pm$0.015}\\
& $\beta$-AnnealVAE~\cite{burgess2018understanding} & 0.911\scalebox{.8}{$\pm$0.002} & - & 0.790\scalebox{.8}{$\pm$0.075} & 0.965\scalebox{.8}{$\pm$0.007} & 0.821\scalebox{.8}{$\pm$0.022} & 0.911\scalebox{.8}{$\pm$0.002} & 0.880\scalebox{.8}{$\pm$0.016}\\
& $\beta$-TCVAE~\cite{chen2018isolating} & 0.914\scalebox{.8}{$\pm$0.008} & - & 0.864\scalebox{.8}{$\pm$0.095} & 0.962\scalebox{.8}{$\pm$0.010} & 0.801\scalebox{.8}{$\pm$0.024} & 0.914\scalebox{.8}{$\pm$0.008} & 0.891\scalebox{.8}{$\pm$0.014}\\
& Factor-VAE~\cite{kim2018disentangling} & 0.916\scalebox{.8}{$\pm$0.004} & - & \textbf{0.893}\scalebox{.8}{$\pm$0.056} & 0.947\scalebox{.8}{$\pm$0.011} & 0.770\scalebox{.8}{$\pm$0.025} & 0.916\scalebox{.8}{$\pm$0.005} & 0.888\scalebox{.8}{$\pm$0.014}\\ \cmidrule(lr){2-9}
& SimCLR~\cite{chen2020simple} & 0.882\scalebox{.8}{$\pm$0.019} & - & 0.767\scalebox{.8}{$\pm$0.025} & 0.976\scalebox{.8}{$\pm$0.011} & 0.863\scalebox{.8}{$\pm$0.036} & 0.876\scalebox{.8}{$\pm$0.015} & 0.873\scalebox{.8}{$\pm$0.016}\\
& \textbf{IP-IRM (Ours)} & \cellcolor{mygray}0.917\scalebox{.8}{$\pm$0.008} & \cellcolor{mygray}- & \cellcolor{mygray} 0.785\scalebox{.8}{$\pm$0.031} & \cellcolor{mygray} \textbf{0.990}\scalebox{.8}{$\pm$0.002} & \cellcolor{mygray} \textbf{0.921}\scalebox{.8}{$\pm$0.009} & \cellcolor{mygray} 0.916\scalebox{.8}{$\pm$0.007} & \cellcolor{mygray} \textbf{0.906}\scalebox{.8}{$\pm$0.011}\\

\hline

\multicolumn{1}{c}{\multirow{7}{*}{\rotatebox{90}{\textbf{Shapes3D}}}} & VAE~\cite{kingma2014ICLR} & 0.351\scalebox{.8}{$\pm$0.026} & 0.284\scalebox{.8}{$\pm$0.009} & \textbf{0.820}\scalebox{.8}{$\pm$0.015} & 0.802\scalebox{.8}{$\pm$0.054} & 0.421\scalebox{.8}{$\pm$0.079} & 0.352\scalebox{.8}{$\pm$0.027} & 0.505\scalebox{.8}{$\pm$0.028}\\
& $\beta$-VAE~\cite{higgins2016beta} & 0.369\scalebox{.8}{$\pm$0.021} & 0.283\scalebox{.8}{$\pm$0.012} & 0.782\scalebox{.8}{$\pm$0.034} & 0.807\scalebox{.8}{$\pm$0.018} & 0.427\scalebox{.8}{$\pm$0.025} & 0.368\scalebox{.8}{$\pm$0.023} & 0.506\scalebox{.8}{$\pm$0.011}\\
& $\beta$-AnnealVAE~\cite{burgess2018understanding} & 0.327\scalebox{.8}{$\pm$0.069} & 0.412\scalebox{.8}{$\pm$0.049} & 0.743\scalebox{.8}{$\pm$0.070} & 0.643\scalebox{.8}{$\pm$0.013} & 0.259\scalebox{.8}{$\pm$0.021} & 0.328\scalebox{.8}{$\pm$0.070} & 0.452\scalebox{.8}{$\pm$0.023}\\
& $\beta$-TCVAE~\cite{chen2018isolating} & 0.470\scalebox{.8}{$\pm$0.035} & 0.291\scalebox{.8}{$\pm$0.023} & 0.777\scalebox{.8}{$\pm$0.031} & 0.821\scalebox{.8}{$\pm$0.054} & 0.439\scalebox{.8}{$\pm$0.084} & 0.469\scalebox{.8}{$\pm$0.034} & 0.545\scalebox{.8}{$\pm$0.032}\\
& Factor-VAE~\cite{kim2018disentangling} & 0.340\scalebox{.8}{$\pm$0.021} & 0.316\scalebox{.8}{$\pm$0.016} & 0.815\scalebox{.8}{$\pm$0.041} & 0.738\scalebox{.8}{$\pm$0.043} & 0.319\scalebox{.8}{$\pm$0.045} & 0.339\scalebox{.8}{$\pm$0.021} & 0.478\scalebox{.8}{$\pm$0.020} \\ \cmidrule(lr){2-9}
& SimCLR~\cite{chen2020simple} & 0.535\scalebox{.8}{$\pm$0.016} & \textbf{0.439}\scalebox{.8}{$\pm$0.030} & 0.678\scalebox{.8}{$\pm$0.050} & 0.949\scalebox{.8}{$\pm$0.005} & 0.733\scalebox{.8}{$\pm$0.055} & 0.536\scalebox{.8}{$\pm$0.015} & 0.645\scalebox{.8}{$\pm$0.026}\\
& \textbf{IP-IRM (Ours)} & \cellcolor{mygray}\textbf{0.565}\scalebox{.8}{$\pm$0.023} & \cellcolor{mygray}0.420\scalebox{.8}{$\pm$0.014} & \cellcolor{mygray} 0.766\scalebox{.8}{$\pm$0.036} & \cellcolor{mygray} \textbf{0.959}\scalebox{.8}{$\pm$0.007} & \cellcolor{mygray} \textbf{0.757}\scalebox{.8}{$\pm$0.025} & \cellcolor{mygray} \textbf{0.565}\scalebox{.8}{$\pm$0.023} & \cellcolor{mygray} \textbf{0.672}\scalebox{.8}{$\pm$0.017}\\

\hline \hline
\end{tabular}}
\caption{Results on disentanglement metrics of existing unsupervised disentanglement methods, standard SSL (SimCLR~\cite{chen2020simple}) and IP-IRM using CMNIST~\cite{arjovsky2019invariant} and Shapes3D~\cite{kim2018disentangling}. Note that IRS is based on intervening the semantics which requires access to the labels of all the semantics, and hence not applicable for CMNIST dataset. Results are averaged over 4 trails (mean $\pm$ std).}
\label{tab:cmnist}
\vspace*{-6mm}
\end{table}

%% file: tsne.tex
\begin{figure}[t!]
    \captionsetup{font=footnotesize,labelfont=footnotesize}
    \centering
    \includegraphics[width=1.0\linewidth]{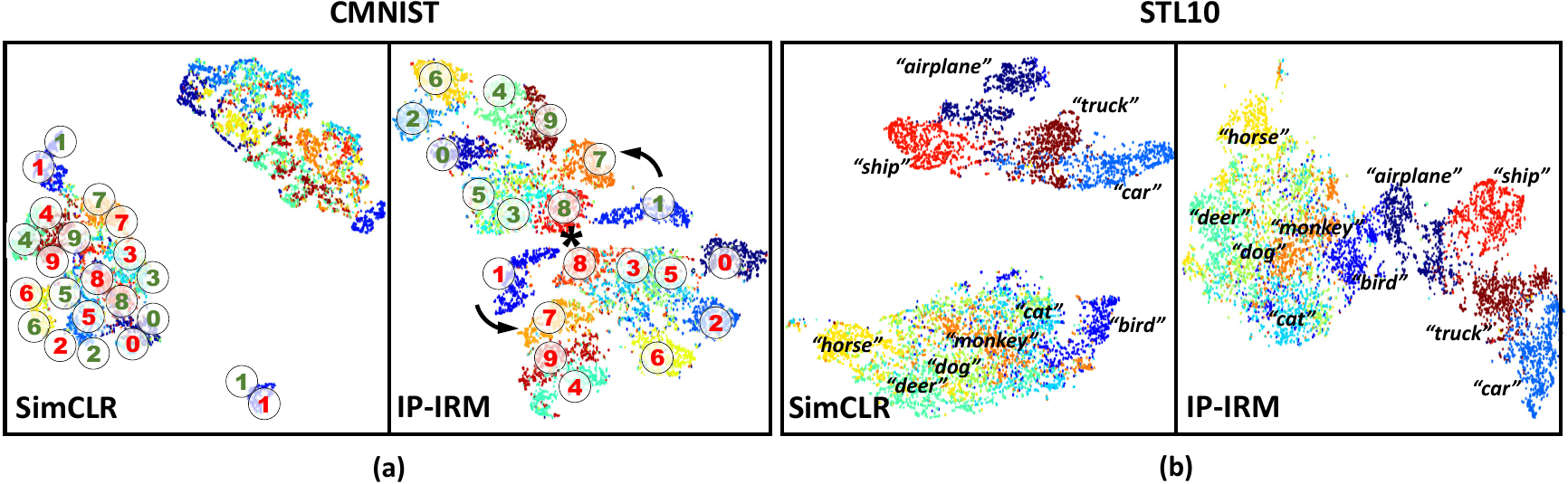}
    \caption{The t-SNE~\cite{van2008visualizing} visualizations of learned feature spaces using SimCLR~\cite{chen2020simple} and IP-IRM on CMNIST~\cite{arjovsky2019invariant} and STL10~\cite{coates2011analysis}. 
    For CMIST in (a), we annotate the digit and color near each cluster. We annotate only half of the feature points for SimCLR to avoid clutter.
    For STL10 in (b), we show the labels of the classes.}
    \label{fig:tsne}
    \vspace{-2mm}
\end{figure}

%% file: dim_group.tex
\begin{figure}[t!]
    \captionsetup{font=footnotesize,labelfont=footnotesize}
    \centering
    \includegraphics[width=1.0\linewidth]{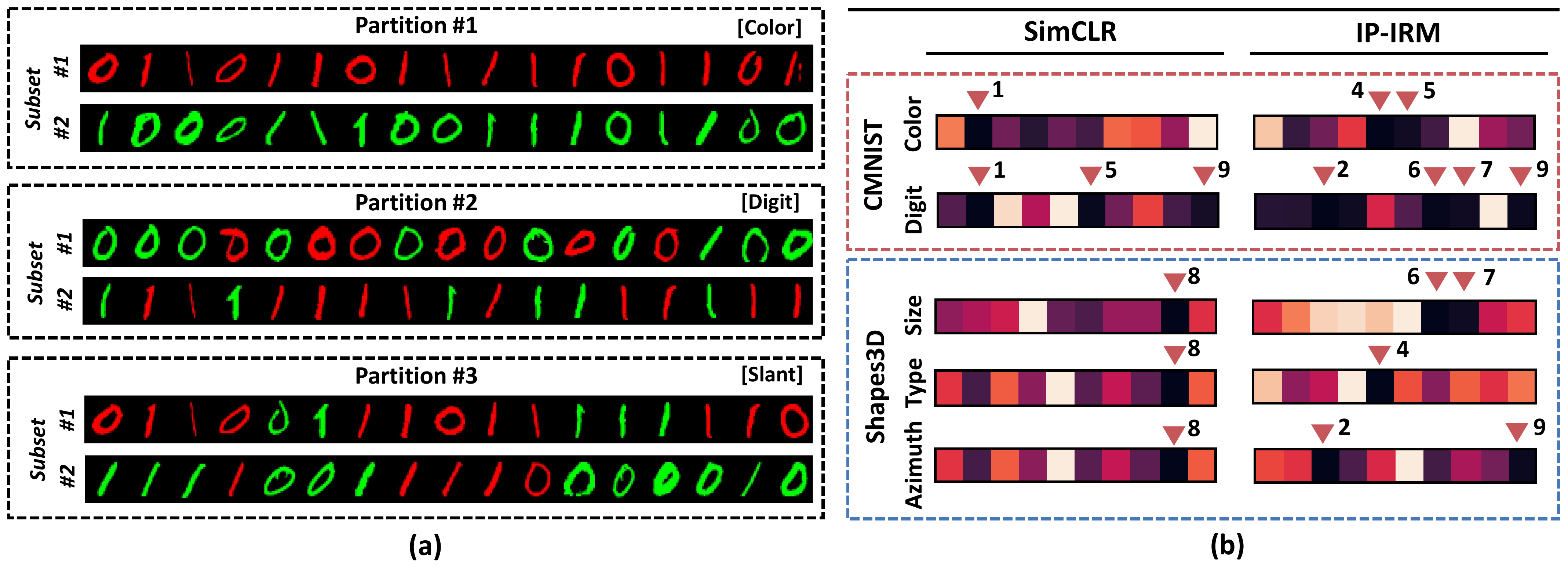}
    \caption{(a) Visualization of the obtained partitions $\mathbf{P}^*$ during training. Each partition has two subset and the displayed images are \emph{randomly} sampled from each subset. (b) Visualization of the variance of each feature dimension when perturbing the semantic indicated on the left. The most equivariant dimensions are indicated by triangles and their corresponding indices.}
    \label{fig:dim_group}
    \vspace{-5mm}
\end{figure}

%% file: 5_Experiments_part2.tex
\subsection{Self-Supervised Learning}
\label{sec:5.2}

\noindent\textbf{Datasets and Settings}.
We conducted the SSL evaluations on 2 standard benchmarks following~\cite{wang2020unsupervised, chuang2020debiased, kalantidis2020hard}.
\textbf{Cifar100}~\cite{krizhevsky2009learning} contains 60,000 images in 100 classes and \textbf{STL10}~\cite{coates2011analysis} has 113,000 images in 10 classes.
We 
used SimCLR~\cite{chen2020simple}, DCL~\cite{chuang2020debiased} and HCL~\cite{kalantidis2020hard} as baselines, and learned the representations for $400$ and $1000$ epochs.
We evaluated both linear and $k$-NN ($k=200$) accuracies for the downstream classification task. Implementation details are in appendix.

\input{ssl_small_modified}

\noindent\textbf{Results}. 
We demonstrate our results and compare with baselines in Table~\ref{tab:ssl_small}.
Incorporating IP-IRM to the 3 baselines brings consistent performance boosts to downstream classification models in all settings, \eg, improving the linear models by 5.55\% on STL10 and 2.92\% on Cifar100.
In particular, we observe that IP-IRM brings huge performance gain with $k$-NN classifiers, \eg, 4.23\% using HCL+IP-IRM on STL10, \ie, the distance metrics in the IP-IRM feature space more faithfully reflects the class semantic differences. This validates that our algorithm further disentangles compared to the standard SSL
Moreover, by extending the training process to 1,000 epochs with MixUp~\cite{lee2021imix}, SimCLR+IP-IRM achieves further performance boost on both datasets, \eg, 5.28\% for $k$-NN and 4.35\% for linear classifier over SimCLR baseline on STL10 dataset. Notably, our SimCLR+IP-IRM surpasses vanilla supervised learning on Cifar100 under the same evaluation setting.
Still, the quality of disentanglement cannot be fully evaluated when the training and test samples are identically distributed---while the improved accuracy demonstrates that IP-IRM representation is more equivariant to class semantics, it does not reveal if the representation is decomposable. Hence we present an out-of-distribution (OOD) setting in Section~\ref{sec:5.3} to further show this property.

\input{figure6}

\noindent\textbf{Is IP-IRM sensitive to the values of hyper-parameters?}
1) \emph{$\lambda_1$ and $\lambda_2$ in Eq.~\eqref{eq:min} and Eq.~\eqref{eq:max}}.
In Figure~\ref{fig:6} (a), we observe that the best performance is achieved with $\lambda_1$ and $\lambda_2$ taking values from $0.2$ to $0.5$ on both datasets. 
All accuracies drop sharply if using $\lambda_1 = 1.0$.
The reason is that a higher $\lambda_1$ forces the model to push the $\phi$-induced similarity to fixed baseline $\theta=1$, rather than decrease the loss $\mathcal{L}$ on the pretext task, leading to poor convergence.
2) \emph{The number of epochs}.
In Figure~\ref{fig:6} (b), we plot the Top-1 accuracies of using $k$-NN classifiers along the 700-epoch training of two kinds of SSL representations---SimCLR and IP-IRM.
It is obvious that IP-IRM converges faster and achieves a higher accuracy than SimCLR.
It is worth to highlight that on the STL10, the accuracy of SimCLR starts to oscillate and grow slowly after the 150-th epoch, while ours keeps on improving.
This is an empirical evidence that IP-IRM keeps on disentangling more and more semantics in the feature space, and has the potential of improvement through long-term training.

\subsection{Potential on Large-Scale Data}

\label{sec:5.3}

\noindent\textbf{Datasets}. We evaluated on the standard benchmark of supervised learning
\textbf{ImageNet ILSVRC-2012}~\cite{deng2009imagenet} which has in total 1,331,167 images in 1,000 classes. To further reveal if a representation is decomposable, we used \textbf{NICO}~\cite{he2021towards}, which is a real-world image dataset designed for OOD evaluations. It contains 25,000 images in 19 classes, with a strong correlation between the foreground and background in the train split (\eg, most dogs on grass).
We also studied the transferability of the
learned representation following
\cite{Ericsson2021HowTransfer, kornblith2019better}: FGVC Aircraft (\textbf{Aircraft})~\cite{maji2013fine}, Caltech-101 (\textbf{Caltech})~\cite{fei2004learning}, Stanford Cars (\textbf{Cars})~\cite{yang2015large}, \textbf{Cifar10}~\cite{krizhevsky2012learning}, \textbf{Cifar100}~\cite{krizhevsky2012learning}, \textbf{DTD}~\cite{cimpoi2014describing}, Oxford 102 Flowers (\textbf{Flowers})~\cite{nilsback2008automated}, Food-101 (\textbf{Food})~\cite{bossard2014food}, Oxford-IIIT Pets (\textbf{Pets})~\cite{parkhi2012cats} and SUN397 (\textbf{SUN})~\cite{xiao2010sun}. 
These datasets include coarse- to fine-grained classification tasks, and vary in the amount of training data (2,000-75,000 images) and classes (10-397 classes), representing a wide range of transfer learning settings.

\noindent\textbf{Settings}. For the ImageNet, all the representations were trained for 200 epochs due to limited computing resources. We followed the common setting~\cite{tian2019contrastive, he2019moco},  using a linear classifier, and report Top-1 classification accuracies. For NICO, we fixed the ImageNet pre-trained ResNet-50 backbone and fine-tuned the classifier. See appendix for more training details.
For the transfer learning, we followed~\cite{Ericsson2021HowTransfer, kornblith2019better} to report the classification accuracies on Cars, Cifar-10, Cifar-100, DTD, Food, SUN and the average per-class accuracies on Aircraft, Caltech, Flowers, Pets. 
We call them uniformly as Accuracy. 
We used the few-shot $n$-way-$k$-shot setting for model evaluation. Specifically, we randomly sampled 2,000 episodes from the \emph{test} splits of above datasets. An episode 
contains $n$ classes, each with $k$ training samples and 15 testing samples, where we fine-tuned the linear classifier (backbone weights frozen) for 100 epochs on the training samples, and evaluated the classifier on the testing samples. We evaluated with $n=k=5$ (results of $n=5,k=20$ in Appendix).

\input{imagenet_nico}

\input{transfer_modified}

\noindent\textbf{ImageNet and NICO}. In Table~\ref{tab:imagenet_nico} ImageNet accuracy, our IP-IRM achieves the best performance over all baseline models.
Yet we believe that this does not show the full potential of IP-IRM, because ImageNet is a larger-scale dataset with many semantics, and it is hard to achieve a full disentanglement of all semantics within the limited 200 epochs.
To evaluate the feature decomposability of IP-IRM, we compared the performance on NICO with various SSL baselines in Table~\ref{tab:imagenet_nico}, where our approach significantly outperforms the baselines by 1.5-4.2\%. This validates IP-IRM feature is more decomposable---if each semantic feature (\eg, background) is decomposed in some fixed dimensions and some classes vary with such semantic, then the classifier will recognize this as a non-discriminative variant feature and hence focus on other more discriminative features (\ie, foreground). In this way, even though some classes are confounded by those non-discriminative features (\eg, most of the ``dog’’ images are with ``grass’’ background), the fixed dimensions still help classifiers neglect those non-discriminative ones. 
We further visualized the CAM~\cite{zhou2016learning} on NICO in Figure~\ref{fig:7}, which indeed shows that IP-IRM helps the classifier focus on the foreground regions.

\noindent\textbf{Few-Shot Tasks.}
As shown in Table~\ref{tab:transfer}, our IP-IRM significantly improves the performance of 5-way-5-shot setting, \eg, we outperform the baseline MoCo-v2 by 2.2\%. This is because IP-IRM can further disentangled $\mathcal{G} / \mathcal{D}_{\textrm{aug}}$ over SSL, which is essential for representations to generalize to different downstream class domains (recall Corollary 2 of Lemma 1).
This is also in line with recent works~\cite{van2019disentangled} showing that a disentangled representation is especially beneficial in low-shot scenarios, and further demonstrates the importance of disentanglement in downstream tasks.

%% file: ssl_small_modified.tex
\begin{wraptable}{l}{0.6\textwidth}
\vspace{-2mm}
\centering
\captionsetup{font=footnotesize,labelfont=footnotesize,skip=5pt}
\setlength\extrarowheight{1pt}
\scalebox{0.85}{
\begin{tabular}{m{3.4cm} p{0.9cm}<{\centering}p{0.9cm}<{\centering}p{0.9cm}<{\centering}p{0.9cm}<{\centering}}
\hline\hline
\multicolumn{1}{c}{\multirow{2}[2]{*}{Method}} &  \multicolumn{2}{c}{\textbf{STL10}} & \multicolumn{2}{c}{\textbf{Cifar100}} \\ \cmidrule(lr){2-3}\cmidrule(lr){4-5}
& $k$-NN & Linear & $k$-NN & Linear\\
\hline
\multicolumn{5}{c}{\underline{\emph{400 epoch training}}} \vspace{3pt} \\
    SimCLR~\cite{chen2020simple} & 73.60 & 78.89 & 54.94 & 66.63\\
    
    DCL~\cite{chuang2020debiased} & 78.82 & 82.56 & 57.29 & 68.59\\
    
    HCL~\cite{kalantidis2020hard} & 80.06 & 87.60 & 59.61 & 69.22\\

    \hline
    
    \textbf{SimCLR+IP-IRM} & \cellcolor{mygray} 79.66 & \cellcolor{mygray} 84.44 & \cellcolor{mygray} 59.10 & \cellcolor{mygray} 69.55\\
    
    \textbf{DCL+IP-IRM} & \cellcolor{mygray} 81.51 & \cellcolor{mygray} 85.36 & \cellcolor{mygray} 58.37  & \cellcolor{mygray} 68.76\\
    
    \textbf{HCL+IP-IRM} & \cellcolor{mygray} \textbf{84.29} & \cellcolor{mygray} \textbf{87.81} & \cellcolor{mygray} \textbf{60.05} & \cellcolor{mygray} \textbf{69.95}\\
    
\hline  \hline
\multicolumn{5}{c}{\underline{\emph{1,000 epoch training}}} \vspace{3pt} \\
SimCLR~\cite{chen2020simple} & 78.60 & 84.24 & 59.45 & 68.73 \\
SimCLR$^{\dagger}$~\cite{lee2021imix} & 79.80 & 85.56 &63.67 &72.18 \\
\textbf{SimCLR$^{\dagger}$+IP-IRM}  &\cellcolor{mygray}\textbf{85.08}  &\cellcolor{mygray}\textbf{89.91}  &\cellcolor{mygray}\textbf{65.82}  &\cellcolor{mygray}\textbf{73.99} \\ \hline
Supervised$^{*}$ & - & - & - & 73.72 \\
Supervised$^{*}$+MixUp~\cite{zhang2018mixup} & - & - &- & 74.19 \\
\hline \hline
\end{tabular}}
\caption{Accuracy (\%) of $k$-NN and linear classifiers on STL10~\cite{coates2011analysis} and Cifar100~\cite{krizhevsky2009learning} using the representations of SimCLR~\cite{chen2020simple}, DCL~\cite{chuang2020debiased}, HCL~\cite{kalantidis2020hard} and those after incorporating our IP-IRM. SimCLR$^{\dagger}$ denotes SimCLR with MixUp regularization. Supervised$^{*}$ represents the supervised learning that keeps the same codebase, optimizer and parameters with SSL stage-2 fine-tuning while only adds the learning rate decay at 60 and 80 epoch.}
\label{tab:ssl_small}
\vspace{-0.4cm}
\end{wraptable}

%% file: figure6.tex
\begin{figure}[t]
\centering
    \captionsetup{font=footnotesize,labelfont=footnotesize}
    \centering
    \includegraphics[width=0.9\linewidth]{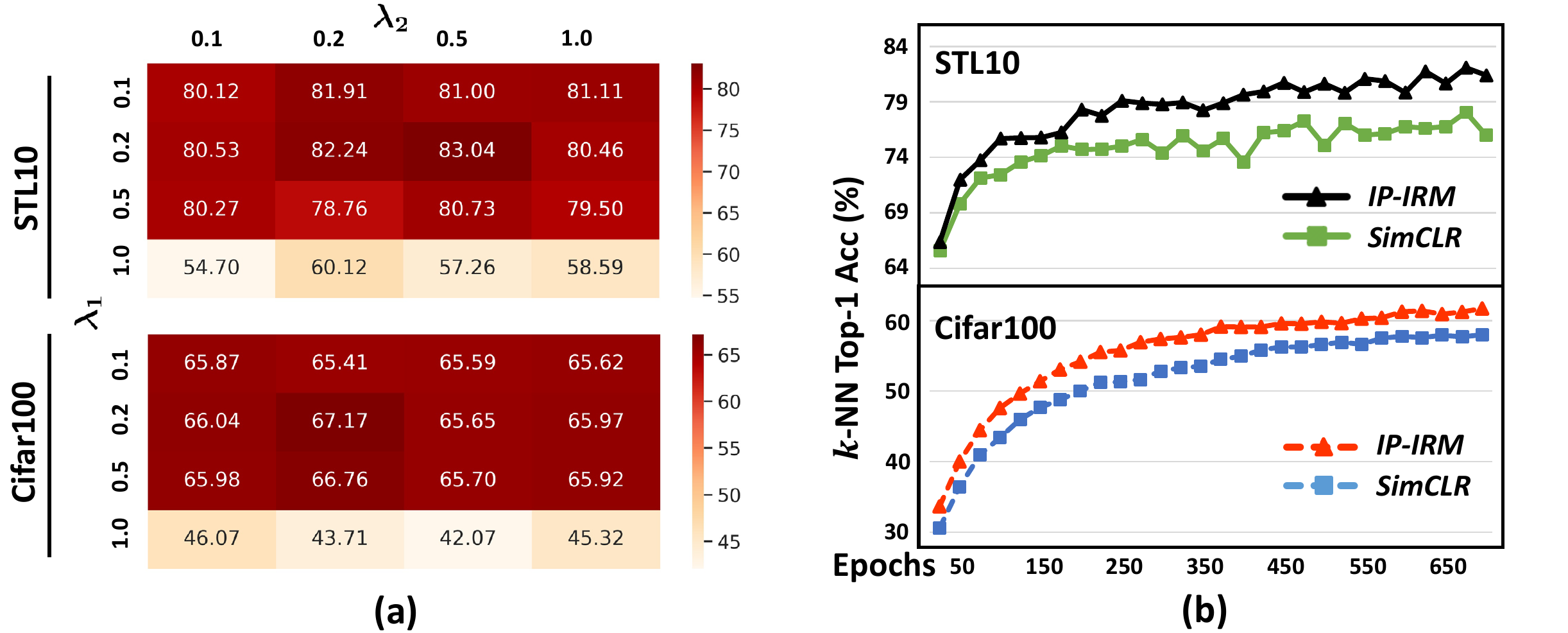}
    \caption{Our ablation study on the STL10 and Cifar100 datasets. 
    (a) The Top-1 accuracy (\%) of linear classifiers using different values of $\lambda_1$ and $\lambda_2$ (in Eq.~\eqref{eq:min} and Eq.~\eqref{eq:max}), by training for 200 epochs on two datasets. 
    (b) The Top-1 accuracy (\%) of $k$-NN classifiers on two datasets, for which we trained the models for 700 epochs and updated $\mathbf{P}$ every 50 epochs.}
    \label{fig:6}
    \vspace{-2mm}
\end{figure}

%% file: imagenet_nico.tex
\begin{figure}[t!]
\centering
\captionsetup{font=footnotesize,labelfont=footnotesize,skip=5pt}
\begin{minipage}[b][][b]{.5\textwidth}
    \centering
    \setlength\tabcolsep{4pt}
    \scalebox{0.85}{
    \begin{tabular}{p{3.5cm} p{1.5cm}<{\centering}p{1cm}<{\centering}}
        \hline\hline
        \multicolumn{1}{c}{Method} & ImageNet & NICO \\
        \hline
        InsDis~\cite{wu2018unsupervised} & 56.5 & 65.6 \\
        PCL~\cite{li2020prototypical} & 61.5   &72.6 \\
        PIRL~\cite{misra2020self} & 63.6   &69.1 \\
        MoCo-v1~\cite{he2019moco} & 60.6 & 69.3 \\
        SimCLR (repro.)~\cite{chen2020simple} & 63.1 & 64.5 \\
        MoCo-v2 (repro.)~\cite{chen2020mocov2} & 67.3 & 78.0 \\
        SimSiam (repro.)~\cite{chen2020exploring} & 68.8 & 66.7 \\ \hline
        \textbf{SimCLR+IP-IRM} & \cellcolor{mygray} 64.8 & \cellcolor{mygray} 66.7 \\
        \textbf{MoCo-v2+IP-IRM} & \cellcolor{mygray} 67.6 & \cellcolor{mygray} \textbf{79.5} \\
        \textbf{SimSiam+IP-IRM} & \cellcolor{mygray} \textbf{69.1} & \cellcolor{mygray}70.9\\
        \hline \hline
    \end{tabular}}
    \captionof{table}{ImageNet and NICO Top-1 Accuracy (\%) of linear classifiers trained on the representations learnt with different SSL methods.}
    \label{tab:imagenet_nico}
\end{minipage}%
\begin{minipage}[b][5cm][b]{.5\textwidth}
    \centering
    \captionsetup{width=0.8\linewidth}
    \includegraphics[width=0.8\linewidth]{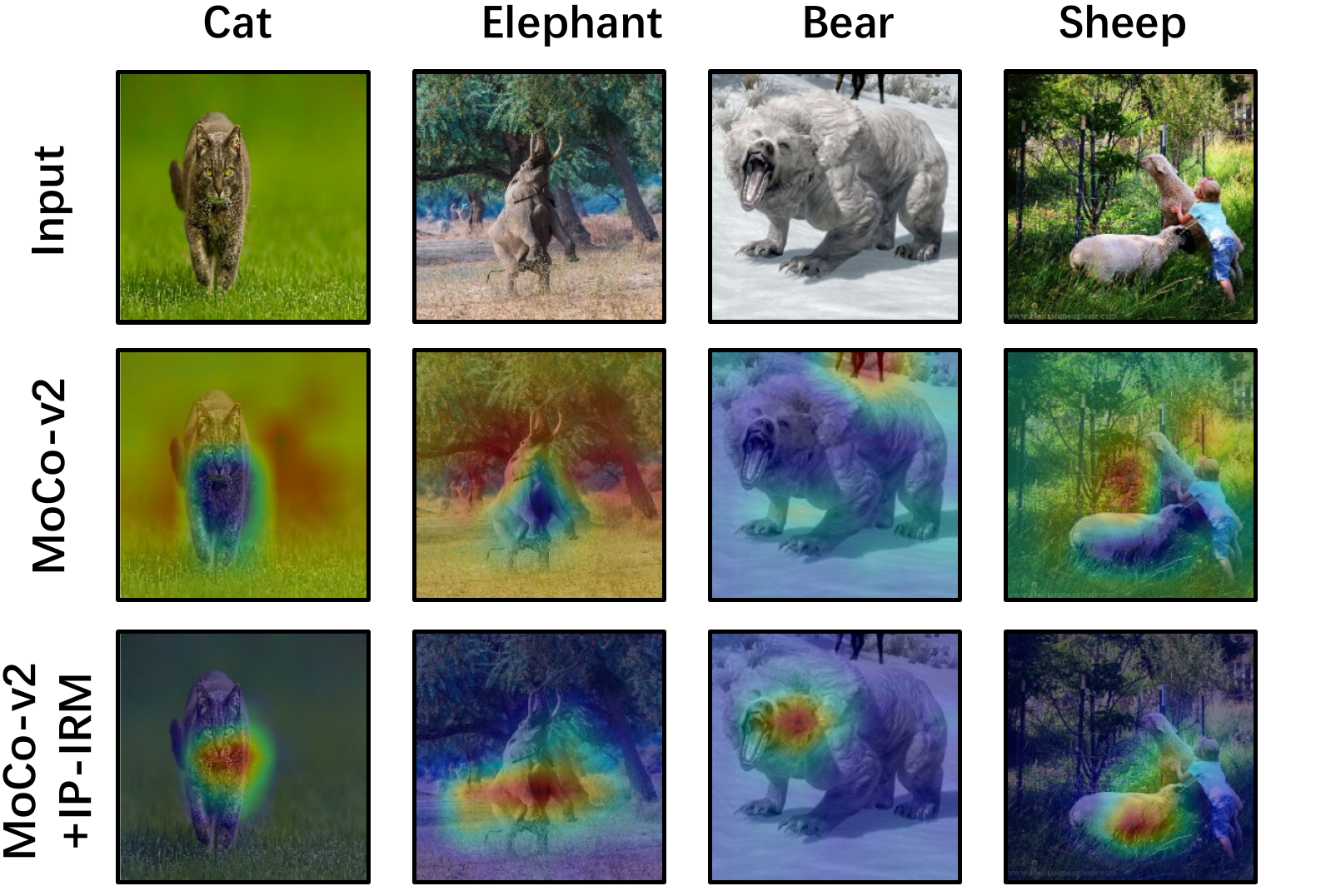}
    \caption{Visualization of CAM~\cite{zhou2016learning} on images from NICO~\cite{he2021towards} dataset using representations of the baseline MoCo-v2~\cite{chen2020mocov2} and our IP-IRM.}
    \label{fig:7}
\end{minipage}
\end{figure}

%% file: transfer_modified.tex
\begin{table}[t!]
\centering
\captionsetup{font=footnotesize,labelfont=footnotesize,skip=5pt}
\setlength\extrarowheight{1pt}
\setlength\tabcolsep{4pt}
\scalebox{0.8}{
\begin{tabular}{p{2.5cm} p{1.0cm}<{\centering}p{1.0cm}<{\centering}p{1.0cm}<{\centering}p{1.0cm}<{\centering}p{1.1cm}<{\centering}p{1.0cm}<{\centering}p{1.0cm}<{\centering}p{1.0cm}<{\centering}p{1.0cm}<{\centering}p{1.0cm}<{\centering}p{1.0cm}<{\centering}}
\hline\hline
Method & Aircraft & Caltech & Cars & Cifar10 & Cifar100 & DTD & Flowers & Food & Pets & SUN & Average\\
\hline

InsDis~\cite{wu2018unsupervised} & 35.07 & 75.97 & 37.49 & 51.49 & 57.61 & 69.38 & 77.35 & 50.01 & 66.38 & 74.97 & 59.57\\
PCL~\cite{li2020prototypical} & \textbf{36.86} & 90.72 & 39.68 & 59.26 & 60.78 & 69.53 & 67.50 & 57.06 & \textbf{88.31} & 84.51 & 65.42\\
PIRL~\cite{misra2020self} & 36.70 & 78.63 & 39.21 & 49.85 & 55.23 & 70.43 & 78.37 & 51.61 & 69.40 & 76.64 & 60.61\\
MoCo-v1~\cite{he2019moco} & 35.31 & 79.60 & 36.35 & 46.96 & 51.62 & 68.76 & 75.42 & 49.77 & 68.32 & 74.77 & 58.69\\
MoCo-v2~\cite{chen2020mocov2} & 31.98 &  92.32 & 41.47 & 56.50 & 63.33 & 78.00 & 80.05 & 57.25 & 83.23 & 88.10  & 67.22\\
\textbf{IP-IRM (Ours)} & \cellcolor{mygray} 32.98 & \cellcolor{mygray} \textbf{93.16} & \cellcolor{mygray} \textbf{42.87} & \cellcolor{mygray} \textbf{60.73} & \cellcolor{mygray} \textbf{68.54} & \cellcolor{mygray} \textbf{79.30} & \cellcolor{mygray} \textbf{82.68} & \cellcolor{mygray} \textbf{59.61} & \cellcolor{mygray} 85.23 & \cellcolor{mygray} \textbf{89.38}  & \cellcolor{mygray} \textbf{69.44}\\

\hline \hline
\end{tabular}}
\caption{Accuracy (\%) of 5-way-5-shot few-shot evaluation using the image representation learned on ImageNet~\cite{deng2009imagenet}. More detailed results are given in Appendix.}
\label{tab:transfer}
\vspace{-6mm}
\end{table}

%% file: 6_Conclusion.tex
\section{Conclusion}

\label{sec:6}

We presented an unsupervised disentangled representation learning method called Iterative Partition-based Invariant Risk Minimization (IP-IRM), based on Self-Supervised Learning (SSL). IP-IRM iteratively partitions the dataset into semantic-related subsets, and learns a representation invariant across the subsets using SSL with an IRM loss. We show that with theoretical guarantee, IP-IRM converges with a disentangled representation under the group-theoretical view, which fundamentally surpasses the capabilities of existing SSL and fully-supervised learning. Our proposed theory is backed by strong empirical results in disentanglement metrics, SSL classification accuracy and transfer performance. IP-IRM achieves disentanglement without using generative models, making it widely applicable on large-scale visual tasks. As future directions, we will continue to explore the application of group theory in representation learning and seek additional forms of inductive bias for faster convergence.

\section*{Acknowledgments and Disclosure of Funding}
The authors would like to thank all reviewers for their constructive suggestions. This research is partly supported by the Alibaba-NTU Joint Research Institute, the A*STAR under its AME YIRG Grant (Project No. A20E6c0101), and the Singapore Ministry of Education (MOE) Academic Research Fund (AcRF) Tier 2 grant.

%% file: appendix/appx_main.tex
\appendix
\noindent\textbf{\huge Appendix}
\vspace{0.5cm}

This is the Appendix for ``Self-Supervised Learning Disentangled Group Representation as Feature''.
Table~\ref{tbl:notation} summarizes the abbreviations and the symbols used in the main paper.

\input{appendix/appx_sections/symbol_table}

This appendix is organized as follows:
\begin{itemize}
\item Section~\ref{sec:pre} provides the preliminary knowledge about the group theory. We also provide a formal definition of ``fixed'' and ``affected'' used in Definition 1.
\item Section~\ref{sec:proof} gives the proofs of Definition 2, Lemma 1 and Theorem 1.
\item Section~\ref{sec:implem} shows the implementation details of our experiments in Section 5 of our main paper.
\item Section~\ref{sec:exp} presents the additional experimental results.
\end{itemize}

\newpage

\input{appendix/appx_sections/preliminary}
\input{appendix/appx_sections/proof}

\input{appendix/appx_sections/implementation}
\input{appendix/appx_sections/exp}
\input{appendix/appx_sections/others}



%% file: appendix/appx_sections/symbol_table.tex

\begin{table}[htbp]
\centering
\begin{center}
\begin{tabular}{lll}
\toprule\toprule
Abbreviation/Symbol & Meaning\\
\midrule
\multicolumn{2}{c}{\underline{\emph{Abbreviation}}} \vspace{3pt} \\
SSL     & Self-supervised Learning\\
SL      & Supervised Learning\\
DCI     & Disentangle Metric for Informativeness\\
IRS     & Interventional Robustness Score\\
EXP     & Explicitness Score \\
MOD     & Modularity Score \\
LR      & Logistic Regression \\
GBT     & Gradient Boosted Trees \\
OOD     & Out-Of-Distributed\\
\midrule
\multicolumn{2}{c}{\underline{\emph{Symbol in Theory}}} \vspace{3pt} \\
$\mathcal{U}$     & Semantic space \\
$\mathcal{X}$     & Vector space\\
$\mathcal{I}$     & Image space\\
$\mathcal{G}$         & Group \\
$\mathcal{G}(\mathbf{x})$              & Group orbit \wrt $\mathcal{G}$ containing the sample $\mathbf{x}$ \\
$\varphi$     & Image generation process $\mathcal{U}\to\mathcal{I}$\\
$\phi$          & Visual representation $\mathcal{I}\to\mathcal{X}$ \\
$f$         & Semantic representation $\mathcal{U}\to\mathcal{X}$\\
$m$         & The number of decomposed subgroups \\

\midrule
\multicolumn{2}{c}{\underline{\emph{Symbol in Algorithm}}} \vspace{3pt} \\
$\mathbf{P}$        & Partition of dataset \\
$\mathbf{P}^*$      & Learned partition through Eq. (3)\\
$\mathcal{P}$       & Set of partitions used in Eq. (2) \\
$N$        & Number of training images \\
$\theta$        & ``Dummy'' parameter used by IRM \\
$I$         & Image \\
$\mathcal{X}_k$       & The set of features in the subset $k$\\
$\mathcal{X}^*$       & The set of augmented view features \\
$\lambda_1$, $\lambda_2$        & Hyper-parameters in Eq. (2) and (3) \\

\bottomrule\bottomrule
\end{tabular}
\end{center}
\caption{List of abbreviations and symbols used in the paper.}
\label{tbl:notation}
\end{table}

%% file: appendix/appx_sections/preliminary.tex
\section{Preliminaries}
\label{sec:pre}

A group is a set together with a binary operation, which takes two elements in the group and maps them to another element. For example, the set of integers is a group under the binary operation of plus. We formalize the notion through the following definition.

\noindent\textbf{Binary Operation}. A binary operation $\cdot$ on a set $\mathcal{S}$ is a function mapping $\mathcal{S} \times \mathcal{S}$ into $\mathcal{S}$. For each $(s_1,s_2)\in \mathcal{S} \times \mathcal{S}$, we denote the element $\cdot(s_1,s_2)$ by $s_1 \cdot s_2$.

\noindent\textbf{Group}. A group $\langle \mathcal{G}, \cdot \rangle$ is a set $\mathcal{G}$, closed under a binary operation $\cdot$, such that the following axioms hold:
\begin{enumerate}[leftmargin=+.2in]
    \item \emph{Associativity}. $\forall g_1, g_2, g_3 \in \mathcal{G}$, we have $(g_1 \cdot g_2) \cdot g_3 = g_1 \cdot (g_2 \cdot g_3)$.
    \item \emph{Identity Element}. $\exists e \in \mathcal{G}$, such that $\forall g\in \mathcal{G}$, $e \cdot g = g \cdot e = g$.
    \item \emph{Inverse}. $\forall g \in \mathcal{G}$, $\exists g' \in \mathcal{G}$, such that $g \cdot g' = g' \cdot g = e$.
\end{enumerate}

Groups often arise as transformations of some space, such as a set, vector space, or topological space. Consider an equilateral triangle. The set of clockwise rotations \wrt its centroid to retain its appearance forms a group $\{60^\circ,120^\circ,180^\circ\}$, with the last element corresponding to an identity mapping. We say this group of rotations act on the triangle, which is formally defined below.

\noindent\textbf{Group Action}. Let $\mathcal{G}$ be a group with binary operation $\cdot$ and $\mathcal{S}$ be a set. An action of $\mathcal{G}$ on $\mathcal{S}$ is a map $\pi:\mathcal{G}\to \mathrm{Hom}(\mathcal{S}, \mathcal{S})$ so that $\pi(e)=\mathrm{id}_\mathcal{S}$ and $\pi(g) \circ \pi(h)=\pi(g\cdot h)$, where $g,h\in \mathcal{G}$, $\circ$ denotes functional composition. $\forall g\in \mathcal{G}, s\in \mathcal{S}$, denote $\pi(g)(s)$ as $g \circ s$.

In our formulation, we have a group $\mathcal{G}$ acting on the semantic space $\mathcal{U}$. For example, consider the color semantic, which can be mapped to a circle representing the hue. Hence the group acting on it corresponds to rotations, similar to the triangle example, \eg, $\pi(g)$ may correspond to rotating a color in $\mathcal{S}$ clockwise by $30^\circ$.
In the context of representation learning, we are interested to learn a feature space to reflect $\mathcal{G}$, formally defined below.

\noindent\textbf{Group Representation}. Let $\mathcal{G}$ be a group. A representation of $\mathcal{G}$ (or $\mathcal{G}$-representation) is a pair $(\pi, \mathcal{X})$, where $\mathcal{X}$ is a vector space and $\pi:\mathcal{G}\to \mathrm{Hom}_{vect}(\mathcal{X},\mathcal{X})$ is a group action, \ie, for each $g\in \mathcal{G}$, $\pi(g):\mathcal{X}\to \mathcal{X}$ is a linear map.

Intuitively, each $g\in\mathcal{G}$ corresponds to a linear map, \ie, a matrix $\mathbf{M}_g$ that transforms a vector $\mathbf{x}\in\mathcal{X}$ to $\mathbf{M}_g\mathbf{x}\in\mathcal{X}$. Finally, there is a decomposition of semantic space and the group acting on it in our definition of disentangled representation. The decomposition of semantic space is based on the Cartesian product $\times$. A similar concept is defined \wrt group.

\noindent\textbf{Direct Product of Group}. Let $\mathcal{G}_1,\ldots,\mathcal{G}_n$ be groups with the binary operation $\cdot$. Let $a_i,b_i\in \mathcal{G}_i$ for $i\in \{1,\ldots,n\}$. Define $(a_1,\ldots,a_n) \cdot (b_1,\ldots,b_n)$ to be the element $(a_1 \cdot b_1,\ldots,a_n \cdot b_n)$. Then $\mathcal{G}_1 \times \ldots \times \mathcal{G}_n$ or $\prod_{i=1}^n \mathcal{G}_i$ is the direct product of the groups $\mathcal{G}_1,\ldots,\mathcal{G}_n$ under the binary operation $\cdot$.

With this, we can formally define \emph{$\mathcal{X}_i,i\in\{1,\ldots,m\}$ is only affected by the action of $\mathcal{G}_i$ and fixed by the action of other subgroups}: $(\pi|\mathcal{G}_j, \mathcal{X}_i)_{j\neq i}$ is a trivial sub-representation (``fixed''), \ie, for each $g\in\mathcal{G}_j,j\neq i$, $\pi(g)$ is the identity mapping $\mathrm{id}_{\mathcal{X}_i}$, and $(\pi|\mathcal{G}_j, \mathcal{X}_i)_i$ is non-trivial (``affected'').

%% file: appendix/appx_sections/proof.tex
\section{Proof}
\label{sec:proof}

\subsection{Proof of Definition 2}

\noindent\textbf{$\mathcal{D}$ Defines a Partition of $\mathcal{X}$}. We will show that $\mathcal{D}$ defines an equivalence relation on $\mathcal{X}$, which naturally leads to a partition of $\mathcal{X}$. For $\mathbf{x}_1,\mathbf{x}_2 \in \mathcal{X}$, let $\mathbf{x}_1 \sim \mathbf{x}_2$ if and only if $\exists g\in \mathcal{D}$ such that $g\cdot \mathbf{x}_1 = \mathbf{x}_2$. We show that $\sim$ satisfies the three properties of equivalence relation. 1) Reflexive: $\forall \mathbf{x} \in \mathcal{X}$, we have $e\cdot \mathbf{x} = \mathbf{x}$, hence $\mathbf{x}\sim \mathbf{x}$. 2) Symmetric: Suppose $\mathbf{x}_1 \sim \mathbf{x}_2$, \ie, $g\cdot \mathbf{x}_1 = \mathbf{x}_2$ for some $g\in \mathcal{D}$. Then $g^{-1} \cdot \mathbf{x}_2 = \mathbf{x}_1$, \ie, $\mathbf{x}_2 \sim \mathbf{x}_1$. 3) Transitive: if $\mathbf{x}_1 \sim \mathbf{x}_2$ and $\mathbf{x}_2 \sim \mathbf{x}_3$, then $g_1 \cdot \mathbf{x}_1 = \mathbf{x}_2$ and $g_2 \cdot \mathbf{x}_2 = \mathbf{x}_3$ for some $g_1,g_2\in \mathcal{D}$. Hence $(g_2 g_1) \cdot \mathbf{x}_1 = \mathbf{x}_3$ and $\mathbf{x}_1 \sim \mathbf{x}_3$.

\noindent\textbf{Number of Orbits}. Recall that $\mathcal{G}$ acts transitively on $\mathcal{X}$ (see Section 4). We consider the non-trivial case where the action of $\mathcal{G}$ is faithful, \ie, the only group element that maps all $\mathbf{x}\in \mathcal{X}$ to itself is the identity element $e$. Let $\mathcal{K}=\mathcal{G} / \mathcal{D}=g_1\times \ldots \times g_k$. We will show that each $c \in \mathcal{K}$ corresponds to a unique orbit. 1) $\forall c\neq e \in \mathcal{K}$, $\mathcal{D}(\mathbf{x}) \neq \mathcal{D}(c \cdot \mathbf{x})$. Suppose $\exists c \neq e \in \mathcal{K}$, such that for some $\mathbf{x} \in \mathcal{X}$, the action of $c$ on each $\mathbf{x}_1\in \mathcal{D}(\mathbf{x})$ corresponds to the identity mapping. One can show that for every different orbit, \ie, $\mathcal{D}(c' \cdot \mathbf{x}) \neq \mathcal{D}(\mathbf{x})$, the action of $c$ on each $\mathbf{x}_2\in \mathcal{D}(\mathbf{x})$ is also identity mapping. As $\mathcal{D}$ partitions $\mathcal{X}$ into orbits \wrt $\mathcal{D}$, this means that the action of $c$ is identity mapping on all $\mathbf{x}\in \mathcal{X}$, which contradicts with the action of $\mathcal{G}$ being faithful. 2) The previous step shows that non-identity group elements in $\mathcal{K}$ lead to a different orbit. We need to further show that these orbits are unique, \ie, $\forall c,c'\neq e \in \mathcal{K}$, if $c\neq c'$, then $\mathcal{D}(c \cdot \mathbf{x}) \neq \mathcal{D}(c' \cdot \mathbf{x})$. Suppose $c' \cdot \mathbf{x} = c \cdot \mathbf{x}$, \ie, $c^{-1} c' \cdot \mathbf{x} = c^{-1} c \cdot \mathbf{x} = \mathbf{x}$, so $c^{-1}c' \in \mathcal{G}_{\mathbf{x}}$, where $\mathcal{G}_{\mathbf{x}}$ is the point stabilizer of $\mathbf{x}$. As the action of $\mathcal{G}$ is faithful, $\mathcal{G}_{\mathbf{x}} = \{e\}$. Hence $c \cdot \mathbf{x} = c' \cdot \mathbf{x}$ implies $c=c'$.

\subsection{Details of Lemma 1}

We will first prove Lemma 1 by showing the representation is $\mathcal{G}$-equivariant, followed by showing that $\mathcal{D}$ and $\mathcal{G} / \mathcal{D}$ are decomposable and finally showing that $\prod_{d\in \mathcal{D}} d$ is not decomposable. We will then present more details on the 4 corollaries.

\noindent\textbf{Proof of $\mathcal{G}$-equivariant}. Suppose that the training loss $-\log \frac{\exp (\mathbf{x}^T_i\mathbf{x}_j)}{\sum\nolimits_{\mathbf{x}\in\mathcal{X}}\exp(\mathbf{x}^T_j\mathbf{x})}$ is minimized, yet $\exists \mathbf{x}_a = \mathbf{x}_b\in\mathcal{X}$ for $a\neq b$. Let $\mathbf{x}_i \in \mathcal{X}$ in the denominator, and we have $\mathbf{x}_j^T \mathbf{x}_i=\mathrm{cos}(\theta_{i,j}) \norm{\mathbf{x}_i} \norm{\mathbf{x}_j}$, where $\theta_{i,j}$ is the angle between the two vectors. When $\mathbf{x}_i = \mathbf{x}_j$, $\mathrm{cos}(\theta_{i,j})=1$. So keeping $\norm{\mathbf{x}_i} \norm{\mathbf{x}_j}$ constant (\ie, the same regularization penalty such as L2), $\mathbf{x}_j^T \mathbf{x}_i$ can be further reduced if $\mathbf{x}_i \neq \mathbf{x}_j$, which reduces the training loss. This contradicts with the earlier assumption. Hence by minimizing the training loss, we can achieve sample-equivariant, \ie, different samples have different features. Note that this does not necessarily mean group-equivariant. However, the variation of training samples is all we know about the group action of $\mathcal{G}$, and we establish that the action of $\mathcal{G}$ is transitive on $\mathcal{X}$, hence we use the sample-equivariant features as the approximation of $\mathcal{G}$-equivariant features.

\noindent\textbf{Proof of Decomposability between $\mathcal{D}$ and $\mathcal{G} / \mathcal{D}$}. Recall the semantic representation $f:\mathcal{U}\to \mathcal{I} \to \mathcal{X}$, which is show to be $\mathcal{G}$-equivariant in the previous step. Consider a non-decomposable representation where $\mathcal{X}$ is affected by the action of both $\mathcal{D}$ and $\mathcal{G} / \mathcal{D}$. Let $\mathcal{X}=\mathcal{X}_d \times \mathcal{X}_c$, where both sub-spaces are affected by the action of the two groups. In particular, denote the semantic representation $f_c:(\mathcal{U}_d \times \mathcal{U}_c) \to \mathcal{X}_c$, where $\mathcal{U}_d$ is affected by the action of $\mathcal{D}$ (recall that $\mathcal{G}$ affects $\mathcal{U}$ and $\mathcal{X}$ through the equivariant map in Figure 1) and $\mathcal{U}_c$ is affected by the action of $\mathcal{G} / \mathcal{D}$. From here, we will construct a representation where $\mathcal{X}_c$ is only affected by the action of $\mathcal{G} / \mathcal{D}$ with a lower training loss.

Specifically, we aim to assign a $d_i^*\in \mathcal{U}_d,i\in\{1,\ldots,k\}$ to the $i$-th orbit, which is given by:
\begin{equation}
    d_1^*,\ldots,d_k^* = \mathop{\mathrm{arg\;min}}_{(d_1,\ldots,d_k)} \mathop{\mathbb{E}}_{i,j\in \{1,\ldots,k\}} f_c(d_i,c_i)^T f_c(d_j, c_j),
    \label{eq:d_gd}
\end{equation}
where $c_i\in \mathcal{U}_c$ is the value of $U_c$ for $i$-th orbit. Now define $f^*_c:\mathcal{U}_c \to \mathcal{X}_c$ given by $f^*_c(c_i) = f_c(d_i^*, c_i) \forall i\in\{1,\ldots,k\}$. Using this new $f^*_c$ has two outcomes:

\noindent 1) $\mathbf{x}^T_i\mathbf{x}_j$ in the numerator is the linear combination of the dot similarity induced from $\mathcal{X}_d$ and $\mathcal{X}_c$. And the dot similarity induced from $\mathcal{X}_c$ is increased, as inside each orbit, the value in $\mathcal{X}_c$ is the same (maximized similarity);

\noindent 2) The denominator is now reduced. This is because the denominator is proportional to $\mathop{\mathbb{E}}_{i,j\in \{1,\ldots,k\}} \mathop{\mathbb{E}}_{d,d'\in \mathcal{U}_d} f_c(d, c_i)^T f_c(d', c_j)$, and we have already selected the best set $d_i^*$ that minimizes the expected dot similarities across orbits.

As the in-orbit dot similarity increases (numerator), and the cross-orbit dot similarity decreases (denominator), the training loss is reduced by decomposing a separate sub-space $\mathcal{X}_c$ affected only by the action of $\mathcal{G} / \mathcal{D}$ with $f^*_c$. Furthermore, note that a linear projector is used in SSL to project the features into lower dimensions, and a linear weight is used in SL. To isolate the effect of $\mathcal{D}$ to maximize the similarity of in-orbit samples (numerator) and exploit the action of $\mathcal{G} / \mathcal{D}$ to minimize the similarity of cross-orbit samples (denominator), the effect of $\mathcal{D}$ and $\mathcal{G} / \mathcal{D}$ on $\mathcal{X}_d$ must be separable by a linear layer, \ie, decomposable. Combined with the earlier proof that $\mathcal{X}_c$ is only affected by the action of $\mathcal{G} / \mathcal{D}$, without loss of generality, we have the decomposition $\mathcal{X}=\mathcal{X}_d \times \mathcal{X}_c$ affected by $\mathcal{D}$ and $\mathcal{G} / \mathcal{D}$, respectively.

\noindent\textbf{Proof of Non-Decomposability of $d\in\mathcal{D}$}. We will show that for a representation with $d\in \mathcal{D}$ decomposed, there exists a non-decomposable representation that achieves the same expected dot similarity, hence having the same contrastive loss. Without loss of generality, consider $d_1, d_2\in \mathcal{D}$ acting on the semantic attribute space $\mathcal{U}_1, \mathcal{U}_2$, respectively. Let $f$ be a decomposable representation such that there exists feature subspaces $\mathcal{X}_1,\mathcal{X}_2\in \mathcal{X}$ affected only by the action of $d_1,d_2$, respectively. Denote $f_1:\mathcal{U}_1\to \mathcal{X}_1, f_2:\mathcal{U}_2\to \mathcal{X}_2$. Now we define a non-decomposable representation with mapping $f'_1:(\mathcal{U}_1,\mathcal{U}_2) \to \mathcal{X}_1$ and $f'_2:(\mathcal{U}_1,\mathcal{U}_2) \to \mathcal{X}_2$, given by $f'_1(U_1=u_1,U_2=u_2)=\frac{1}{\sqrt{2}} (f_1(U_1=u_1) + f_2(U_2=u_2))$ and $f'_2(U_1=u_1,U_2=u_2)=\frac{1}{\sqrt{2}} (f_1(U_1=u_1) - f_2(U_2=u_2))$. Now for any pair of samples with semantic $(u_1,u_2)$ and $(u'_1,u'_2)$, $u_1,u'_1 \in \mathcal{U}_1$ and $u_2,u'_2\in \mathcal{U}_2$, the dot similarity induced from the subspace $\mathcal{X}_1,\mathcal{X}_2$ is given by $f_1(u_1)^2 + f_2(u_2)^2$. Therefore, the decomposed and non-decomposed representations yield the same expected dot similarity over all pairs of samples, and have the same contrastive loss.

\noindent\textbf{Corollary 1}. This follows immediately from the proof on $\mathcal{G}$-equivariance.

\noindent\textbf{Corollary 2}. The same proof above holds for the SL case with $\mathbf{x}_i \in \mathcal{X}$, where $\mathcal{X}$ is the set of classifier weights. In this view, each sample in the class can be seen as an augmented view (augmented by shared attributes such as view angle, pose, etc) of the class prototype. In downstream learning, the \emph{shared} attributes are not discriminative, hence the performance is affected mostly by $\mathcal{G} / \mathcal{D}$. For example, if the groups corresponding to ``species'' and ``shape'' act on the same feature subspace (entangled), such that ``species''=``bird'' always have ``shape''=``streamlined'' feature, this representation does not generalize to downstream tasks of classifying birds without streamlined shape (\eg, ``kiwi'').

\noindent\textbf{Corollary 3}. In SL and SSL, the model essentially receives supervision on attributes that are not discriminative towards downstream tasks, through augmentations and in-class variations, respectively. The group $\mathcal{D}$ acts on the semantic space of these attributes, hence $|\mathcal{D}|$ determines the amount of supervision received. With a large $|\mathcal{D}|$, the model filters off more irrelevant semantics and $\mathcal{G} / \mathcal{D}$ more accurately describe the differences between classes. Note that the standard image augmentations in SSL are also used in SL, making $|\mathcal{D}|$ even larger in SL.

\noindent\textbf{Corollary 4}. When the number of samples in some orbit(s) is smaller than $|\mathcal{D}(\mathbf{x})|$, this has two consequences that prevent disentanglement: 1) The $\mathcal{G}$-equivariance is not guaranteed as the training samples do not fully describe $\mathcal{G}$. 2) The decomposability is not guaranteed as the decomposed $f^*_c, f^*_d$ in the previous proof only generalizes to the seen combination of the value in $\mathcal{U}_d \times \mathcal{U}_c$.

\subsection{Proof of Theorem 1}
\label{sec:b3}

We will first revisit the Invariant Risk Minimization (IRM). Let $\mathcal{I}$ be the image space, $\mathcal{X}$ the feature space, $\mathcal{Y}$ the classification output space (\eg, the set of all probabilities of belonging to each class), the feature extractor backbone $\phi:\mathcal{I}\to \mathcal{X}$ and the classifier $\omega:\mathcal{X}\to \mathcal{Y}$. Let $\mathcal{E}_{tr}$ be a set of training environments, where each $e \in \mathcal{E}_{tr}$ is a set of images. IRM aims to solve the following optimization problem:
\begin{equation}
\begin{split}
    &\mathop{\mathrm{min}}_{\phi,\omega} \sum_{e\in \mathcal{E}_{tr}} R^e(\omega \circ \phi)\\
    \mathrm{subject\;to}\; \omega &\in \mathop{\mathrm{arg\,min}}_{\bar{\omega}} R^e(\bar{\omega}\circ \phi)\;\; \forall e\in\mathcal{E}_{tr},
    \label{eq:irm}
\end{split}
\end{equation}
where $R^e(\omega\circ\phi)$ is the empirical classification risk in the environment $e$ using backbone $\phi$ and classifier $\omega$. Conceptually, IRM aims to find a representation $\phi$ such that the optimal classifier on top of $\phi$ is the same for all environments. As Eq.~\eqref{eq:irm} is a challenging, bi-leveled optimization problem, it is initiated into the practical version:
\begin{equation}
    \mathop{\mathrm{min}}_{\phi} \sum_{e\in \mathcal{E}_{tr}} R^e(\phi) + \lambda \norm{\nabla_{\omega=1.0} R^e(\omega \cdot \phi)}^2,
\end{equation}
where $\lambda$ is the regularizer balancing between the ERM term and invariant term.

The above IRM is formulated for supervised training. In SSL, there is no classifier mapping from $\mathcal{X}\to \mathcal{Y}$. Instead, there is a projector network $\sigma:\mathcal{X} \to \mathcal{Z}$ mapping features to another feature space $\mathcal{Z}$, and Eq. (1) is used to compute the similarity with positive key (numerator) and negative keys (denominator) in $\mathcal{Z}$. Note that $h$ in SSL is not equivalent to $\phi$ in SL, as $\sigma$ itself does not generate the probability output like $\omega$, rather, the comparison between positive and negative keys does.

In fact, the formulation of contrastive IRM is given by Corollary 2 of Lemma 1, which says that SL is a special case of contrastive learning, and the set of all classifier weights is the positive and negative key space. In IRM with SL, we are trying to find a set of weights $\omega_{\mathrm{SL}}$ from the classifier weights space (\eg, $\mathbb{R}^{d\times c}$ with feature dimension as $d$ and number of classes as $c$) that achieves invariant prediction. Hence in IRM with SSL, we are trying to find a set of keys $\omega_{\mathrm{SSL}}$ from the key space (\eg, $\mathbb{R}^{z\times n}$ with $z$ being the dimension of $\mathcal{Z}$ and number of positive and negative keys as $n$) that achieves invariant prediction by differentiating a sample with negative keys (Note that the similarity with positive keys is maximized and fixed using standard SSL training by decomposing augmentations and other semantics as in Lemma 1). Specifically, in IP-IRM, the 2 subsets in each partition form the set of training environments $\mathcal{E}_{tr}$.

\noindent\textbf{Proof of the Sufficient Condition}. Suppose that the representation is fully disentangled \wrt $\mathcal{G} / \mathcal{D}_{\textrm{aug}}$. By Definition 1, there exists subspace $\mathcal{X}_i\in \mathcal{X}$ affected only by the action of $c_i \in \mathcal{G} / \mathcal{D}_{\textrm{aug}}$. For each partition given by $\{\mathcal{G}'(c_i\cdot \mathbf{x}), \mathcal{G}'(c_i^{-1}\cdot \mathbf{x})\}$, let the projector network $\sigma^*:\mathcal{X}' \to \mathcal{Z}$, where $\mathcal{X}'=\mathcal{X}_1 \times \ldots \times \mathcal{X}_{i-1} \times \mathcal{X}_{i+1} \times \ldots \times \mathcal{X}_k$. Note that $\sigma^*$ can be achieved in the parameter space of a linear layer, as the subspace $\mathcal{X}_i$ is decomposed into fixed dimensions for all samples, which can be filtered out by a linear layer by setting weights associated to those dimensions as $0$. Moreover, the resulting space $\mathcal{Z}$ (\ie, space of positive and negative keys) is affected only by the action of $\mathcal{G}'$. As the in-orbit group corresponds to $\mathcal{G}'$, the values in $\mathcal{X}_i$ are not discriminative towards SSL objective. Hence there exists $\sigma^*$ mapping to $\mathcal{Z}$ that is optimal in both orbits, \ie, minimizing the contrastive IRM loss.

\noindent\textbf{Proof of the Necessary Condition}. Suppose that the contrastive IRM loss is minimized for the partition $\{\mathcal{G}'(c_i\cdot \mathbf{x}), \mathcal{G}'(c_i^{-1}\cdot \mathbf{x})\}$. We will show that $c_i$ is disentangled. First we will consider the space $\mathcal{Z}$ (mapped from $\mathcal{X}$ by the projector network). If $\mathcal{Z}$ is affected by the action of $c_i$ and $\mathcal{G}'$, let $\omega^*_1\in \mathop{\mathrm{arg\,min}}_{\bar{\omega}} R^1(\bar{\omega}\circ \phi)$ that optimizes the contrastive loss in the first orbit (by exploiting the equivariance of $\mathcal{G}'$). Given a fixed $\phi$, $\omega^*_1$ is unique as the contrastive loss is convex. The IRM constraint requires that $\omega^*_1 = \omega^*_2$, which means that the action of $c_i$ on $\mathcal{Z}$ corresponds to identity mapping. Yet this contradicts with the equivariant property from Corollary 1 of Lemma 1. Therefore when the contrastive IRM loss is minimized, $\mathcal{Z}$ cannot be affected by the action of $c_i$. Given the linear projector network, there are two possibilities for $\mathcal{X}$: 1) $\mathcal{X}=\mathcal{X}'_i \times \mathcal{X}'$ where $\mathcal{X}'_i$ is affected by the action of $c_i$ and $\mathcal{G}$', and $\mathcal{X}'$ is affected by the action of $\mathcal{G}'$. In this way, the projector $\sigma$ can discard $\mathcal{X}'_i$ to obtain $\mathcal{Z}$ unaffected by $c_i$. However, under the representation $\phi$ leading to the feature space with decomposition $\mathcal{X}=\mathcal{X}'_i \times \mathcal{X}'$, the optimal $\omega$ in each orbit will exploit $\mathcal{X}'_i$, which is discriminative as it is affected by $\mathcal{G}'$. Therefore, there exists no $\omega$ that is \emph{simultaneously} optimal between the two orbits under this representation. 2) $\mathcal{X}=\mathcal{X}_i \times \mathcal{X}'$ where $\mathcal{X}_i$ is affected only by the action of $c_i$, and $\mathcal{X}'$ is affected by the action of $\mathcal{G}'$, \ie, the representation is disentangled with $c_i$. In this way, we have $\omega$ that is simultaneously optimal across orbits as shown in the proof of the sufficient condition.

\noindent\textbf{Example of Step 2}. As we consider two orbits in each partition corresponding to $g_i$, we denote $\mathcal{U}_i \in \{0,1\} \forall i \in \{1,\ldots,k\}$ as a binary attribute space affected by $g_i$. We will show how maximizing the contrastive IRM loss will lead to the partition where cross-orbit group element $h\in\mathcal{G} / \mathcal{D}$.

From Lemma 1, the representation $f$ is equivariant under the action of $\mathcal{G}$, \ie, the SSL loss $\mathcal{L}$ reveal the information about $g\in \mathcal{G}$. Specifically, each pair of samples in a subset corresponds to a group element whose action transform the attribute of the first sample to the second sample. In a subset, if more group elements corresponding to pair-wise transformation are identity mapping, the samples in the group are more similar in semantics. With the equivariant property of the representation, the features in the subset are also more similar, leading to larger $\mathcal{L}$ (more difficult to distinguish two samples apart). 

Given binary semantic attributes, the similarity in the semantics is measured by the hamming distance, \ie, for $u_1,u_2\in \prod_{i=1}^{k} \mathcal{U}_i$, their hamming distance $d_H(u_1,u_2)$ is given by the number of different bits (\eg, $d_H(u_1=01,u_2=11)=1$). Denote the set of semantic attributes of the samples in the two orbits in space $\mathcal{U}_1,\ldots,\mathcal{U}_{k}$ as $\mathcal{S}_1$ and $\mathcal{S}_2$, respectively. From the equivariant property of Lemma 1, the first term in Eq. (3) $\mathcal{L} (\phi,\theta=1.0,k=1,\mathbf{P}) + \mathcal{L} (\phi,\theta=1.0,k=2,\mathbf{P})$ is maximized when the average hamming distance in the two orbits $d(\mathcal{S}_1) + d(\mathcal{S}_2)$ is minimized, where $d(\mathcal{S})$ is given by:
\begin{equation}
    d(\mathcal{S})=\frac{1}{|\mathcal{S}|^2}\sum_{u\in \mathcal{S}} \sum_{u' \in \mathcal{S}} d_H(u,u').
    \label{eq:max_goal}
\end{equation}

\begin{figure}[h!]
\captionsetup{font=footnotesize,labelfont=footnotesize}
    \centering
    \includegraphics[width=.88\linewidth]{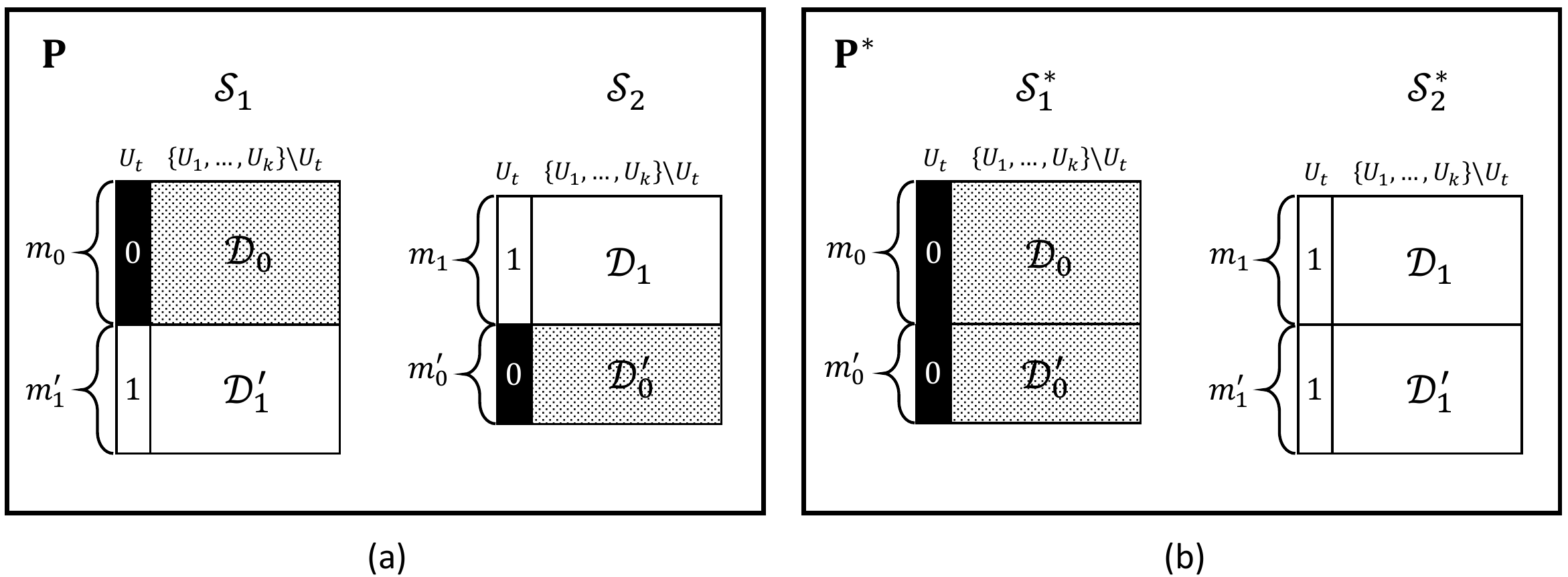}
    \caption{(a) Any arbitrary data partition $\mathbf{P}$. (b) The partition $\mathbf{P}^*$.}
    \label{fig:max_irm}
\end{figure}

Without loss of generality, an arbitrary data partition is illustrated in Figure~\ref{fig:max_irm} (a), where we arrange the order of the samples in each orbit, such that those with $U_{t}=0,t\in\{1,\ldots,k\}$ come first in the subset $\mathcal{S}_1$, and those with $U_{t}=1$ come first in the subset $\mathcal{S}_2$. The subset $\mathcal{S}_1$ has $m_0$ samples with $U_{t}=0$ and $m'_1$ samples with $U_{t}=1$. Denote $\mathcal{D}_0=\{(u_1,\ldots,u_{t-1},u_{t+1},\ldots,u_{k}) \mid u\in \mathcal{S}_1 \land u_{t}=0\}$, $\mathcal{D}'_1=\{(u_1,\ldots,u_{t-1},u_{t+1},\ldots,u_{k}) \mid u\in \mathcal{S}_1 \land u_{t}=1\}$. In orbit $\mathcal{S}_2$, we define $m_1,m'_0,\mathcal{D}_1,\mathcal{D}'_0$ similar to in $\mathcal{S}_1$. In Figure~\ref{fig:max_irm} (b), we show the partition $\mathbf{P}^*$ as described in Theorem 1. We will proceed to show $\forall \mathcal{S}_1,\mathcal{S}_2$:
\begin{equation}
    d(\mathcal{S}_1) + d(\mathcal{S}_2) \geq d(\mathcal{S}_1^*) + d(\mathcal{S}^*_2).
\end{equation}

As the hamming distance is calculated per-dimension, we have the following decomposition:
\begin{equation}
\begin{split}
    d(\mathcal{S}_1)&=d(\mathcal{D}_0 \cup \mathcal{D}'_1) + \frac{1}{|\mathcal{S}_1|^2}\sum_{u\in \mathcal{S}_1} \sum_{u' \in \mathcal{S}_1} d_H(u_{t+1},u'_{t+1})\\
    &= d(\mathcal{D}_0 \cup \mathcal{D}'_1) + \frac{2m_0 m'_1}{(m_0+m'_1)^2}
\end{split}
\end{equation}
\begin{equation}
\begin{split}
    d(\mathcal{S}_2)&=d(\mathcal{D}_1 \cup \mathcal{D}'_0) + \frac{1}{|\mathcal{S}_2|^2}\sum_{u\in \mathcal{S}_2} \sum_{u' \in \mathcal{S}_2} d_H(u_{t+1},u'_{t+1})\\
    &= d(\mathcal{D}_1 \cup \mathcal{D}'_0) + \frac{2m_1 m'_0}{(m_1+m'_0)^2}
\end{split}
\end{equation}
\begin{equation}
    d(\mathcal{S}^*_1)=d(\mathcal{D}_0 \cup \mathcal{D}'_0)
\end{equation}
\begin{equation}
    d(\mathcal{S}^*_2)=d(\mathcal{D}_1 \cup \mathcal{D}'_1)
\end{equation}

We will prove Eq.~\eqref{eq:max_goal} by induction. First consider the case where $|\mathcal{D}'_1|=|\mathcal{D}'_0|=1$. Denote $\mathcal{D}'_1=\{d'_1\}$ and $\mathcal{D}'_0=\{d'_0\}$. We can expand $d(\mathcal{S}_1)$ as
\begin{equation}
    d(\mathcal{S}_1) = d(\mathcal{D}_0) + \frac{2}{(m_0 + 1)^2} \sum_{d_0\in \mathcal{D}_0} d_H(d_0, d'_1).
\end{equation}
We can similarly expand $d(\mathcal{S}_2),d(\mathcal{S}^*_1)$ and $d(\mathcal{S}^*_2)$. Once the same terms are cancelled out, to prove Eq.~\ref{eq:max_goal}, we only need to show for any $d'_0,d'_1$, we have:
\begin{equation}
    m_0 + \sum_{d_0\in \mathcal{D}_0} d_H(d_0, d'_1) + m_1 + \sum_{d_1\in \mathcal{D}_1} d_H(d_1, d'_0) \geq \sum_{d_0\in \mathcal{D}_0} d_H(d_0, d'_0) + \sum_{d_1\in \mathcal{D}_1} d_H(d_1, d'_1),
    \label{eq:hamming_criteria}
\end{equation}
One sufficient condition is that the number of elements in $\mathcal{D}_0,\mathcal{D}_1$ is $2^{t-1}$. This can be empirically achieved with a large dataset. First, we will prove:
\begin{equation}
    m_0 + \sum_{d_0\in \mathcal{D}_0} d_H(d_0, d'_1) \geq \sum_{d_0\in \mathcal{D}_0} d_H(d_0, d'_0).
\end{equation}
Consider $\mathcal{D}_0$ with $2^{t-1}$ unique elements. If the elements in $\mathcal{D}_0$ are such that they lie in a sub-cube of dimension $t-1$ with $U_j=a$ for $j\in \{1,\ldots,t\},a\in\{0,1\}$. Let
\begin{equation}
    \Delta = \sum_{d_0\in \mathcal{D}_0} d_H(d_0, d'_0) - \sum_{d_0\in \mathcal{D}_0} d_H(d_0, d'_1)
\end{equation}

\begin{wrapfigure}{r}{0.24\textwidth}
    \captionsetup{font=footnotesize,labelfont=footnotesize}
    \centering
    \raisebox{0pt}[\dimexpr\height-0.6\baselineskip\relax]{\includegraphics[width=1.0\linewidth]{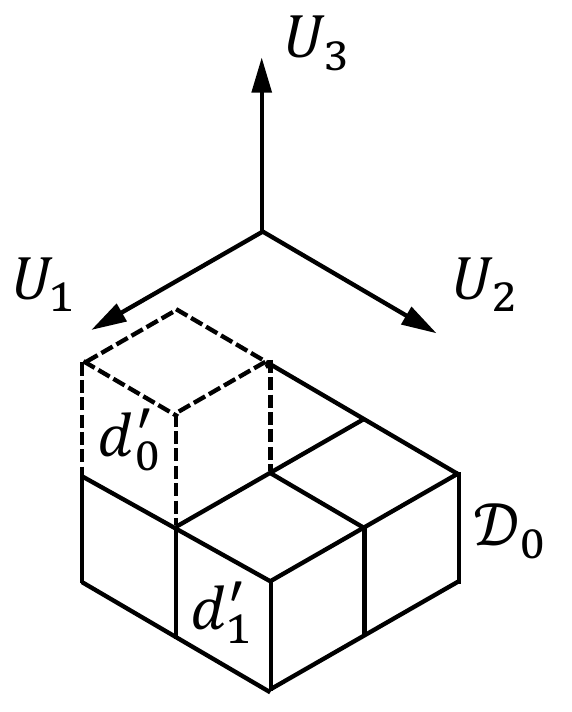}}
    \caption{An example of the sub-cube case with $m_0=4$, $t=3$ and $U_3=0$ in $\mathcal{D}_0$. The maximum $\Delta$ is 4 with $U_3=1$ for $d'_0$ and $U_3=0$ for $d'_1$.}
    \label{fig:subcube}
    \vspace{-6mm}
\end{wrapfigure}
It is easy to show that $\Delta$ is maximized when $d'_0 \not\in \mathcal{D}_0$ and $d'_1 \in \mathcal{D}_0$, where $\Delta=m_0$ (example in Figure~\ref{fig:subcube}). This satisfies Eq.~\eqref{eq:hamming_criteria}. Now consider a general case of $\mathcal{D}_0$ such that $n_a$ elements have $U_j=a$, forming the set $\mathcal{D}_{0,a}$ and $n_{\bar{a}}$ elements have $U_j=\bar{a}$, forming the set $\mathcal{D}_{0,\bar{a}}$, where $\bar{(\cdot)}$ denotes negation. Without loss of generality, let $n_a > n_{\bar{a}}$ (we discuss equality case later). To maximize $\Delta$, $d'_0$ must have $U_j=\bar{a}$, which can be proved through contradiction. If $U_j=a$ for $d'_0$ that maximizes $\Delta$, there exists $d''_0$ which differs from $d'_0$ only on $U_j$, such that $d_H(d''_0,d)-d_H(d'_0,d)=1 \;\forall d\in \mathcal{D}_{0,a}$ and $d_H(d''_0,\bar{d})-d_H(d'_0,\bar{d})=1 \;\forall \bar{d}\in \mathcal{D}_{0,\bar{a}}$. As $n_a>n_{\bar{a}}$, we have $\sum_{d_0\in \mathcal{D}_0} d_H(d_0, d'_0) < \sum_{d_0\in \mathcal{D}_0} d_H(d_0, d''_0)$, which contradicts the condition that we begin with. One can similarly show that $d'_1$ must have $U_j=a$. In the case of equality, both $d'_0,d'_1$ can have $U_j=a$ or $U_j=\bar{a}$. Now, starting with the case where every element in $\mathcal{D}_0$ has $U_j=a$ (sub-cube case), for every one additional element in $\mathcal{D}_0$ that we change its $U_j$ to $\bar{a}$ which $n_a>=n_{\bar{a}}$ still holds, $\sum_{d_0\in \mathcal{D}_0} d_H(d_0, d'_0)$ is reduced by 1 and $\sum_{d_0\in \mathcal{D}_0} d_H(d_0, d'_1)$ is increased by 1, which reduces $\Delta$ by 2. Hence $\Delta$ is maximized in the sub-cube case, for which we have already shown Eq.~\eqref{eq:hamming_criteria} holds, and we have proven the sufficient condition.

One can apply the same analysis above and prove:
\begin{equation}
    m_1 + \sum_{d_1\in \mathcal{D}_1} d_H(d_1, d'_0) \geq \sum_{d_1\in \mathcal{D}_1} d_H(d_1, d'_1).
\end{equation}



Now we have proved the case where $|\mathcal{D}'_0|=|\mathcal{D}'_1|=1$. By induction, we assume that Eq.~\eqref{eq:max_goal} holds for $|\mathcal{D}'_0|=|\mathcal{D}'_1|=p, p\geq 1$. We need to prove for the case with one additional element, \ie, $\mathcal{D}'_1 \cup d''_1$ and $\mathcal{D}'_0 \cup d''_0$. Through reduction, this is to show:
\begin{equation}
    m_0 + \sum_{d_0\in \mathcal{D}_0} d_H(d_0, d''_1) + m_1 + \sum_{d_1\in \mathcal{D}_1} d_H(d_1, d'_0) \geq \sum_{d_0\in \mathcal{D}_0} d_H(d_0, d''_0) + \sum_{d_1\in \mathcal{D}_1} d_H(d_1, d'_1).
\end{equation}
This clearly holds given Eq.~\eqref{eq:hamming_criteria} is true. 

Eq.~\eqref{eq:max_goal} contains an equality case, we have shown that in the sufficient condition that the equality holds in the sub-cube case, \ie, the partition is such that within every subset, the group action of $c_i,i\in\{1,\ldots,k\}$ on the subspace of $\mathcal{X}$ spanned by the subset is the identity mapping. This means that with the overall SSL loss term itself, the partition may be based on one of $c_1,\ldots,c_k$. However, with the contra-position of the sufficient condition in Theorem 1, the cross-orbit action does not correspond to any $d\in \mathcal{D}$. Hence overall, we have shown the maximization leads to a partition with cross-orbit action as $h\in\mathcal{G} / \mathcal{D}$.

%% file: appendix/appx_sections/implementation.tex
\section{Implementation Details}
\label{sec:implem}

\subsection{Implementation Details of the CNN Activation Visualization}

We used ImageNet100 with image size of 224 for our visualization in Figure 2 (b). The visualization is based on the guided propagation method used in~\cite{springenberg2014striving}, and we adopted the publicly available implementation\footnote{\url{https://github.com/utkuozbulak/pytorch-cnn-visualizations}}. We chose VGG-16~\cite{karen2015very} as backbone, due to its native support by guided propagation visualization. 
We followed the default training methods of the SimCLR but only replaced the default ResNet-50 backbone with the VGG16. We trained the baseline and ours model with 200 epochs and for IP-IRM, $\lambda_1=0.2$, $\lambda_2=0.5$.
Please refer to Section~\ref{sec:ssl_imp} for more details. 
Once the backbone is trained, we first obtained the augmentation-unrelated filters by performing augmentations and removing the filters equivariant to augmentations.
Then we performed K-Means clustering ($K=4$) on the CNN weights of layer 17 and layer 28 among the remaining filters, and chose the 4 filters closest to the cluster center. The motivation for such design is to reveal what the representation captures beyond augmentation-equivariant semantics, and the clustering helps locate the different semantics captured by the representation. We show the activation visualization on layer 18 and 29, \ie, after the ReLU layer, which is the actual input to the next CNN layer, instead of immediately after CNN layer 17 and 28.

\subsection{Unsupervised Disentanglement}

\subsubsection{Evaluation Metric Details}

As we discussed in Section~5.1 of the main paper, here we follow~\cite{locatello2019challenging, zaidi2020measuring} to give more detailed formulas or evaluation methods of the used metrics. All the implementations follow the open-source library~\footnote{\url{https://github.com/google-research/disentanglement_lib}}.

\noindent\textbf{Disentangle Metric for Informativeness (DCI)}.
In~\cite{eastwood2018framework}, the authors propose to unify three properties of the representations, \ie, modularity, compactness and explicitness into a complete framework instead of a single metric. It computes the importance of each dimension of the learned feature for predicting a factor of variation (semantic attribute). The predictive importance $R_{ij}$ of the dimensions of feature can be computed with a Lasso or a Random Forest classifier. For the lasso regressor, the importance weights $R_{ij}$ are the magnitudes of the weights learned by the model, while the Gini importance~\cite{breiman2001random} of feature dimensions is used with random forests.
Here we adopted the Informativeness score, which can be computed as the prediction error of predicting the factors of variations.

\noindent\textbf{Interventional Robustness Score (IRS)}.
IRS~\cite{suter2019robustly} provides a causal view for the disentangled generative process and further introduces a new metric to evaluate the interventional effects on the feature representations. 
The key idea behind IRS is that traversing nuisance factors should not impact feature dimensions corresponding to the targeted factors, \eg, changing the color of the banana should not affect its shape.
Specifically, when evaluating the factor $f_i$, a reference set is created which contains instances with the same factor $f_i$ but different nuisance factors. Then the distance is computed between the mean of feature dimensions corresponding to the target factor $f_i$. This computation process is repeated for different nuisance factors and the maximum distance will be reported as the worst case. The maximum distances of each factor are weighted averaged by the frequency of the factor realizations in the data set to obtain the final IRS score.

\noindent\textbf{Modularity Score (MOD) \& Explicitness Score (EXP)}.
The modularity score measures the mutual information (MI) between each feature dimension and each semantic attribute~\cite{ridgeway2018learning}.
In the ideal situation (largest MOD score), a single feature dimension should have high MI with a single attribute and zero MI with all other attributes.
Given a feature dimension index $i$ and the index of attribute $f$, we denote the MI between the feature and factor by $m_{i,f} \in \mathbb{R}$. For faster computation, a vector $\mathbf{t}_i$ is created with the same size as the semantic attribute, which represents the best-matching case of ideal modularity for code dimension $i$:
$$
t_{i,f}=\left\{\begin{array}{ll}
\theta_{i}, & \text { if } f=\arg \max _{g}\left(m_{i,g}\right) \\
0, & \text { otherwise,}
\end{array}\right.
$$
where $\theta_{i}=\max _{g}\left(m_{i,g}\right)$. The observed deviation from the template is given by
$$
\delta_{i}=\frac{\sum_{f}\left(m_{i,f}-t_{i,f}\right)^{2}}{\theta_{i}^{2}(N-1)},
$$
where $N$ is the number of attributes. A deviation of 0 indicates that the perfect MOD score and 1 indicates that this dimension has equal MI with every attribute. Therefore, $1-\delta_{i}$ is used as a MOD score for feature dimension $i$ and the mean of $1-\delta_{i}$ over $i$ as the MOD score for the overall feature.
For explicitness (EXP), a classifier is trained on the entire feature for predicting semantic factor by assuming that attributes have discrete values. Specifically, a one-versus-rest logistic regression classifier can be trained to predict the attributes, and the ROC area-under-the-curve (AUC) is reported as the final explicitness score.

\noindent\textbf{Downstream Tasks with LR and GBT}.
We follow~\cite{locatello2019challenging} to consider the simplest downstream classification task. Its target is to predict the true factors of variations from the learned feature using either multi-class logistic regression (LR) or gradient boosted trees (GBT). 
Specifically, for each factor a simple model is fitted for prediction and then we report the average accuracy across factors on test set. Following~\cite{locatello2019challenging}, we adopt two different models. First, we train a cross validated logistic regression (LR) with the regularization strength as 10 ($Cs = 10$) and folds as 5 using Scikit-learn package. Then, we train a gradient boosting classifier (GBT) with default parameters. We sample the training set of 7500 and the evaluation set of 2500 samples for CMNIST; while the training set of 1000 and the evaluation set of 2500 for Shapes3D.

\subsubsection{Model Architecture}
In the main paper we stated that we used CNN-based feature extractor bascknones with comparable number of parameters for all the baselines and IP-IRM. Here we provide the detailed model architectures in Table~\ref{table:vae_cmnist},~\ref{table:vae_shape3d},~\ref{table:ssl_cmnist},~\ref{table:ssl_shape3d}. 
\input{appendix/tables/model_disentangle}

\subsubsection{Training Details}

For all the methods, the training epoch is fixed to 200 and training batch size is fixed to 2048. The optimizer is Adam and feature dimension was set to 10.
Note that due to much more computational complexity of the self-supervised learning, we \emph{decreased} the sample size used for training and evaluation with both VAE and SSL models to make sure a fair comparison. Specifically, for CMNIST, we used 50,000 samples for training and 10,000 for testing. For Shapes3D, we used 90,000 for training and 10,000 for testing.
For VAE-based model, we followed their common hyperparameters setting without fine-tuning and the implementation codes were built upon the open-source code~\footnote{\url{https://github.com/YannDubs/disentangling-vae}}.
Specifically, the learning rate is 5e-4 for all the models except for the 1e-4 of Factor-VAE.
For $\beta$-VAE, $\beta=4$. 
For $\beta$-AnnealVAE,  the starting annealed capacity $C=0$ and the final annealed capacity $C=25$.
For $\beta$-TCVAE, $\beta=6, \alpha=1, \gamma=1$.
For Factor-VAE, $\gamma=6$, the learning rate of the discriminator is 5e-5 with Adam optimizer.
For SSL-based methods, we used SimCLR as our base model. The learning rate is set to 1e-3 and temperature is 0.5. For IP-IRM, $\lambda_1=0.2$, $\lambda_2=0.5$, the partition $\mathbf{P}$ is updated every 30 epochs and optimized from random every time.
All the experiments were completed on the work station with 4 Nvidia 2080Ti GPUs.

\subsection{Self-supervised Learning}

\subsubsection{Implementation Details}
\label{sec:ssl_imp}

For SimCLR, we follow~\cite{chuang2020debiased, kalantidis2020hard} to use ResNet-50 as the encoder architecture and use the Adam optimizer. The temperature is set to 0.5 and the dimension of the latent vector is 128. All the models were trained for 400 or 1000 epochs and evaluated by training a linear classifier after fixing the learned features. The learning rate is set to 0.001 and weight decay is 1e-6 for both SSL pretraining and downstream fine-tuning.
For DCL~\cite{chuang2020debiased} and HCL~\cite{kalantidis2020hard}, we adopted the best parameters posted in the original paper and followed the implementation from the open-source code~\footnote{\url{https://github.com/chingyaoc/DCL}, \url{https://github.com/joshr17/HCL}}. Particularly, for DCL, $\tau^{+}=0.12$ on STL10 and Cifar100 dataset; while for HCL, $\tau^{+}=0.1, \beta=1$ on STL10,  $\tau^{+}=0.05, \beta=0.5$ on Cifar100. 
For our IP-IRM, we applied $\lambda_1=0.2, \lambda_2=0.5$ on STL10 and $\lambda_1=0.2, \lambda_2=0.2$ on Cifar100. 
We additionaly performed baselines and our IP-IRM on ImageNet-100~\cite{tian2019contrastive}, a randomly chosen subset of 100 classes of ImageNet with 200 training epochs. The results are reported in Table~\ref{tab:appx_ssl_small}. We followed the best parameters reported in the paper~\cite{chuang2020debiased, kalantidis2020hard}. Particularly, for DCL, $\tau^{+}=0.01$. For HCL, $\tau^{+}=0.01, \beta=1.0$. 
Note that on ImageNet100, we slightly modified the $\tau$ and $\beta$ to achieve the better performance. Specifically, for DCL+IP-IRM, we set $\tau=0.1$; while for HCL+IP-IRM, we set $\tau=0.1$, $\beta=0.5$.
While extending ssl pretraining process to 1000 epochs with MixUp~\cite{lee2021imix}, we directly followed the open-source MixUp implementation\footnote{\url{https://github.com/kibok90/imix}} for SimCLR: the temperature is set to 0.2 and MixUp alpha is set to 1.0. For our IP-IRM, note that the MixUp was processed within each subsets to avoid the sample confusion of two subsets. Therefore, the training time inevitably grow linearly due to the increase of the partitions. To mitigate this problem, we controlled the number of partitions (\eg, 5 partitions, FIFO) in practice as an approximation.
For the supervised training, as introduced in the main paper, we adopted the same codebase, optimizer and parameter setting, \ie, the Adam optimizer with learning rate as 0.001 and weight decay as 1e-6 for 100 epochs. We only added the learning rate decay to achieve the optimal training at 60 and 80 epoch. Moreover, we found that adding MixUp with $\alpha=1.0$ would decrease the performance. Therefore, we set $\alpha$ as 0.5.
The experiments were completed on the workstation with 4 Nvidia 2080Ti GPUs.

When training on the large-scale ImageNet, we built our IP-IRM based on SimCLR, MoCo-v2 and SimSiam~\cite{chen2020exploring}~\footnote{\url{https://github.com/taoyang1122/pytorch-SimSiam} (Note that the official implementation of SimSiam was not available then.)}. The batch size is 512 for SimCLR due to the limited computation resources. Different with the contrastive loss adopted in SimCLR and other conventional SSL methods, SimSiam discards the negative samples and uses the MSE loss. Therefore, our IP-IRM was built directly based on MSE loss and encourages the samples in each subset achieve the same performance (\ie, the same MSE loss). For the detailed proof that IRM can be applied to any convex loss function, please refer to~\cite{arjovsky2019invariant}.
$\lambda_1=0.2, \lambda_2=0.5$, partition $\mathbf{P}$ was updated every 50 epochs. Other hyper-parameters followed the default SimSiam setting.
For downstream linear classifier training, we followed the open-source code to use Nvidia LARC optimizer with learning rate 1.6. The experiments were completed on the workstation with 8 Nvidia V100 GPUs.

\subsection{OOD Classification on NICO}

\input{appendix/tables/nico_stastic}

\subsubsection{NICO dataset}
In our experiment, we selected a subset of NICO animal dataset~\cite{he2021towards} as a challenging benchmark to test the feature decomposability for proposed IP-IRM and baselines. 
Specifically, images in NICO are labeled with a context background class (\eg, ``on grass''), besides the object foreground class (\eg, ``dog''). For each animal class, we randomly sample its images and make sure the context labels of those images are within a fixed set of 10 classes (e.g.,``snow'', ``on grass'' and ``in water''). 
Based on these data, we propose a challenging OOD setting including three factors regarding contexts: 
1) Long-Tailed: training context labels are in long-tailed distribution in each individual class, e.g., ``sheep'' might have 10 images of ``on grass'', 5 images of ``in water'' and 1 image of ``on road'';
2) Zero-Shot: for each object class, 7 out of 10 context labels are in training images and the other 3 labels appear only in testing;
3) Orthogonal—the head context label of each object class is set to be as unique (dominating only in one object class) as possible.
\input{appendix/figures/ir_cropped}
The detailed 7 context classes (long-tailed contexts) and 3 zero-shot contexts are shown in Table~\ref{tab:nico}) for each object class. 
Next, we formed a long-tailed training dataset by selecting part of the images in each context class with multiplying a ratio. 
In particular, the ratio for $w$-th context class ($w \in \{0,\ldots,6\}$) is given by
\begin{equation}
    \text{ratio} = \mathrm{IR}^{w / 6},
\end{equation}
where $\mathrm{IR}$ is a hyper-parameter that denotes the imbalance ratio. The effect of $\mathrm{IR}$ on ratio is shown in Figure~\ref{fig:ratio} --- lower ratio leads to the harder OOD problem. In the main paper we keep $\mathrm{IR}=0.02$.
During testing, the number of test samples across the 7 context classes are balanced, \ie, 50 samples per context. Moreover, we added 3 zero-shot context classes for each object class as shown in Table~\ref{tab:nico} (last three columns). These zero-shot context classes have the larger number of test samples (100 samples per context). Therefore, a model that performs well in our split must be robust to both long-tailed and zero-shot problems \textit{w.r.t.} the context class. Figure~\ref{fig:nico_dataset} shows an example of our constructed subset for ``cat'' and ``dog'' during training and testing.

\input{appendix/figures/nico_dataset}

\subsubsection{Training Details}

Different from the linear classifier fine-tuning of SSL which uses a linear fc layer mapping from the feature dimension to the class label dimension, we adopted a linear fc layer mapping from the feature dimension to the feature dimension (\eg, $\mathbb{R}^{2048\times 2048}$) as the feature space mapping for bias NICO training. Then the classification (both training and inference) is based on the metric learning paradigm (\eg, supervised contrastive learning) by measuring the distance between sample features (\eg, $k$-nn accuracy). The reason is that the shared feature mapping layer can help classifier to neglect the bias feature. Empirically, this paradigm outperforms the conventional classifier by more than 10\%.
For parameter setting details, the learning rate is set to 0.2 with SGD optimizer and the batch size is fixed to 128. We utilized $k$-nn classifier ($k=10$) for evaluation. All the models are trained for 150 epochs and the first 2 epochs are the warm-up stage. Learning rate was decreased by 5 at 80, 120 epoch. We report the best accuracy during training as the final performance.

\subsection{Transfer Learning}
We used the ResNet-50 backbone trained on ImageNet through SSL or our IP-IRM as the feature extractor when transferring to the downstream tasks.
For the baseline models, we followed~\cite{Ericsson2021HowTransfer} to download the pre-trained weights of the ResNet50 models in the open-source code.
All models have 23.5M parameters in their backbones and were pre-trained on the ImageNet~\cite{deng2009imagenet} training set, consisting of 1.28M images.

\input{appendix/figures/visualize_group_full_balance}

\subsubsection{Many-shot Learning}
The top-1 accuracy metric is reported on Food-101, CIFAR-10, CIFAR-100, SUN397, Stanford Cars, and DTD, mean per-class accuracy on FGVC Aircraft, Oxford-IIIT Pets, Caltech-101, and Oxford 102 Flowers and the 11-point mAP metric on Pascal VOC 2007. On Caltech-101 we randomly selected 30 images per class to form the training set and we test on the rest. We used the first train/test split defined in DTD and SUN397. On FGVC Aircraft, Pascal VOC2007, DTD, and Oxford 102 Flowers we used the validation sets defined by the authors, and on the other datasets we randomly select 20\% of the training set to form the validation set. The optimal hyperparameters were selected on the validation set, after which we retrained the model on all training and validation images. Finally, the accuracy is computed on the test set.
For the downstream transfer learning, we finetuned the models following~\cite{Ericsson2021HowTransfer} with minor modifications. We train for 5000 steps
with a batch size of 64. The optimiser is SGD with Nesterov momentum and a momentum parameter of 0.9. The learning rate follows a cosine annealing schedule without restarts, and the initial learning rate is chosen from a grid of 4 logarithmically spaced values between 0.0001 and 0.1. 
\input{appendix/tables/ssl_imagenet100}
The weight decay is similarly chosen from a grid of 4 logarithmically spaced values between 1e-6 and 1e-3, along with no weight decay. These weight decay values are divided by the learning rate. We selected the data augmentation from: random crop with resize and flip, or simply a center crop.

\subsubsection{Few-shot Learning}
For the dataloader, no augmentation is used and the images are resized to 224 pixels along the shorter side using bicubic resampling, followed by a center crop of 224 $\times$ 224.
For all methods on all datasets, we trained a linear classifier in each episode using SGD optimizer with learning rate as 0.01 and weight decay as 0.001. The classifier was trained for 100 epochs with batch size as 4. The learned classifier is evaluated using 15 query images in each episode and the reported accuracies and errors are computed on 2000 total episodes.

%% file: appendix/tables/model_disentangle.tex
\begin{table*}[h!]
\captionsetup{font=footnotesize,labelfont=footnotesize}
\centering
\vspace{2mm}
\begin{tabular}{l  l}
\toprule
\textbf{Encoder} & \textbf{Decoder}\\
\midrule 
Input: $28\times 28 \times 3$                       & Input: $\mathbb{R}^{10}$\\
FC, $2352\times 148$ ReLU             & FC, $10\times 148$ ReLU\\
FC, $148\times 148$ ReLU             & FC, $148\times 148$ ReLU\\
FC, $148\times 20$             & FC, $148\times 2352$ Sigmoid\\
\bottomrule
\end{tabular}
\caption{Encoder and Decoder architecture of the VAE-based methods with 0.747M parameters for CMNIST dataset in the main experiment.}
\label{table:vae_cmnist}
\end{table*}

\begin{table*}[h!]
\captionsetup{font=footnotesize,labelfont=footnotesize}
\centering
\vspace{2mm}
\begin{tabular}{l  l}
\toprule
\textbf{Encoder} & \textbf{Decoder}\\
\midrule 
Input: $64\times 64 \times 3$                       & Input: $\mathbb{R}^{10}$\\
$4\times 4$ conv, 32 ReLU, stride 2, padding 1             & FC, $10\times 32$ ReLU\\
$4\times 4$ conv, 32 ReLU, stride 2, padding 1             & FC, $32\times 512$ ReLU\\
$4\times 4$ conv, 32 ReLU, stride 2, padding 1             & $4\times 4$ upconv, 32 ReLU, stride 2, padding 1\\
$4\times 4$ conv, 32 ReLU, stride 2, padding 1             & $4\times 4$ upconv, 32 ReLU, stride 2, padding 1    \\
FC, $512\times 32$ ReLU                                     & $4\times 4$ upconv, 32 ReLU, stride 2, padding 1 \\
FC, $32\times 20$                               & $4\times 4$ upconv, 3 Sigmoid, stride 2, padding 1 \\
\bottomrule
\end{tabular}
\caption{Encoder and Decoder architecture of the VAE-based methods with 0.136M parameters for Shapes3D dataset in the main experiment.}
\label{table:vae_shape3d}
\end{table*}

\begin{table*}[h!]
\captionsetup{font=footnotesize,labelfont=footnotesize}
\centering
\vspace{2mm}
\begin{tabular}{l}
\toprule
\textbf{Encoder} \\
\midrule
Input: $28\times 28 \times 3$               \\
FC, $2352\times 256$ ReLU           \\
FC, $256\times 256$ ReLU            \\
FC, $256\times 10$   \\
\midrule
\textbf{Projection Head} \\
\midrule
FC, $10\times 64$ BatchNorm ReLU        \\
FC, $64\times 32$ \\

\bottomrule
\end{tabular}
\caption{Model architecture of the SSL models with 0.674M parameters for CMNIST dataset in the main experiment.}
\label{table:ssl_cmnist}
\end{table*}

\begin{table*}[h!]
\captionsetup{font=footnotesize,labelfont=footnotesize}
\centering
\vspace{2mm}
\begin{tabular}{l}
\toprule
\textbf{Encoder} \\
\midrule
Input: $64\times 64 \times 3$               \\
$4\times 4$ conv, 32 BatchNorm ReLU, stride 2, padding 1           \\
$4\times 4$ conv, 32 BatchNorm ReLU, stride 2, padding 1            \\
$4\times 4$ conv, 64 BatchNorm ReLU, stride 2, padding 1  \\
$4\times 4$ conv, 64 BatchNorm ReLU, stride 2, padding 1  \\
$4\times 4$ Average Pooling \\
FC, $64\times 10$   \\
\midrule
\textbf{Projection Head} \\
\midrule
FC, $10\times 64$ BatchNorm ReLU        \\
FC, $64\times 32$ \\

\bottomrule
\end{tabular}
\caption{Model architecture of the SSL models with 0.120M parameters for Shapes3D dataset in the main experiment.}
\label{table:ssl_shape3d}
\end{table*}

%% file: appendix/tables/nico_stastic.tex
\begin{table*}[t]
\captionsetup{font=footnotesize,labelfont=footnotesize}
\centering
\scalebox{0.65}{ 
\setlength{\tabcolsep}{1.7mm}{
    \begin{tabular}{ccccccccccc}
\toprule\toprule
\diagbox{\textbf{Class}}{\textbf{Context}}    & \multicolumn{7}{c}{\large{Long-Tailed Contexts}}                                                & \multicolumn{3}{c}{\large{Zero-Shot Contexts}}     \\ \cmidrule(lr){1-1}\cmidrule(lr){2-8}\cmidrule(lr){9-11}
Dog      & on grass  & in water     & in cage   & eating       & on beach  & lying    & running   & at home      & in street    & on snow     \\ \hline
Cat      & on snow   & at home      & in street & walking      & in river  & in cage  & eating    & in water     & on grass     & on tree     \\ \hline
Bear     & in forest & black        & brown     & eating grass & in water  & lying    & on snow   & on ground    & on tree      & white       \\ \hline
Sheep    & eating    & on road      & walking   & on snow      & on grass  & lying    & in forest & aside people & in water     & at sunset   \\ \hline
Bird     & on ground & in hand      & on branch & flying       & eating    & on grass & standing  & in water     & in cage      & on shoulder \\ \hline
Rat      & at home   & in hole      & in cage   & in forest    & in water  & on grass & eating    & lying        & on snow      & running     \\ \hline
Horse    & on beach  & aside people & running   & lying        & on grass  & on snow  & in forest & at home      & in river     & in street   \\ \hline
Elephant & in zoo    & in circus    & in forest & in river     & eating    & standing & on grass  & in street    & lying        & on snow     \\ \hline
Cow      & in river  & lying        & standing  & eating       & in forest & on grass & on snow   & at home      & aside people & spotted     \\ \hline
Monkey   & sitting   & walking      & in water  & on snow      & in forest & eating   & on grass  & in cage      & on beach     & climbing    \\ \bottomrule\bottomrule
\end{tabular}}}
\vspace{0.2cm}
\caption{Construction of our NICO~\cite{he2021towards} subset for OOD multi-classification . \textbf{Context} denotes the context class name, while \textbf{Class} represents the object class name. ``Long-Tailed Contexts'' is the training contexts arranged by the sample number order (from more to less) and ``Zero-shot Contexts'' represents the context labels only appear in testing rather than training.
}
\label{tab:nico}
\end{table*}

%% file: appendix/figures/ir_cropped.tex
\begin{wrapfigure}{r}{0.5\textwidth}
\captionsetup{font=footnotesize,labelfont=footnotesize}
\centering
\includegraphics[width=0.95\linewidth]{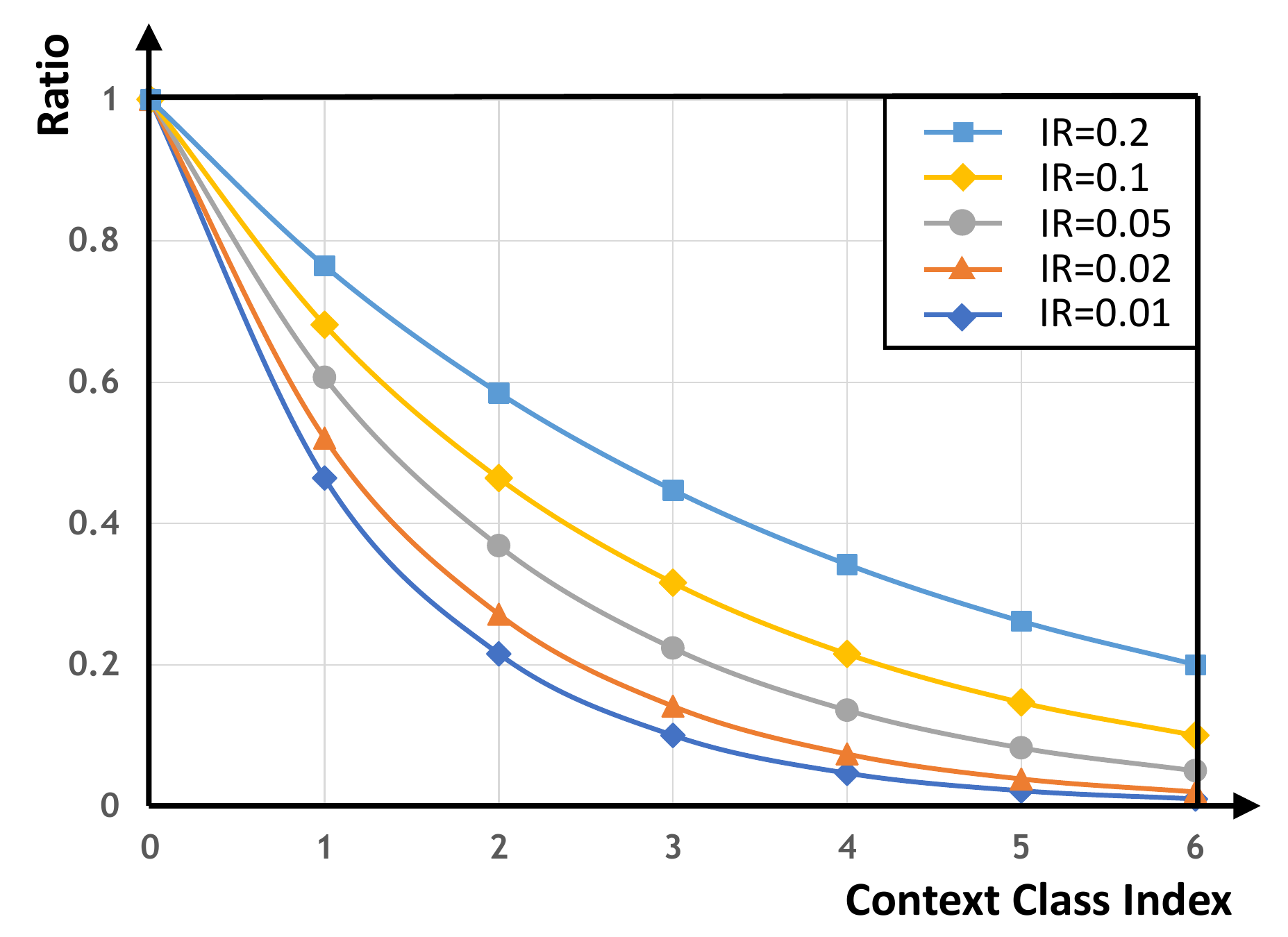}
\caption{Plot of context class index against its corresponding ratio under various imbalance ratio (IR).}
\label{fig:ratio}
\vspace{0.2cm}
\end{wrapfigure}

%% file: appendix/figures/nico_dataset.tex
\begin{figure*}[h]
\captionsetup{font=footnotesize,labelfont=footnotesize}
\begin{center}
\includegraphics[width=0.95\textwidth]{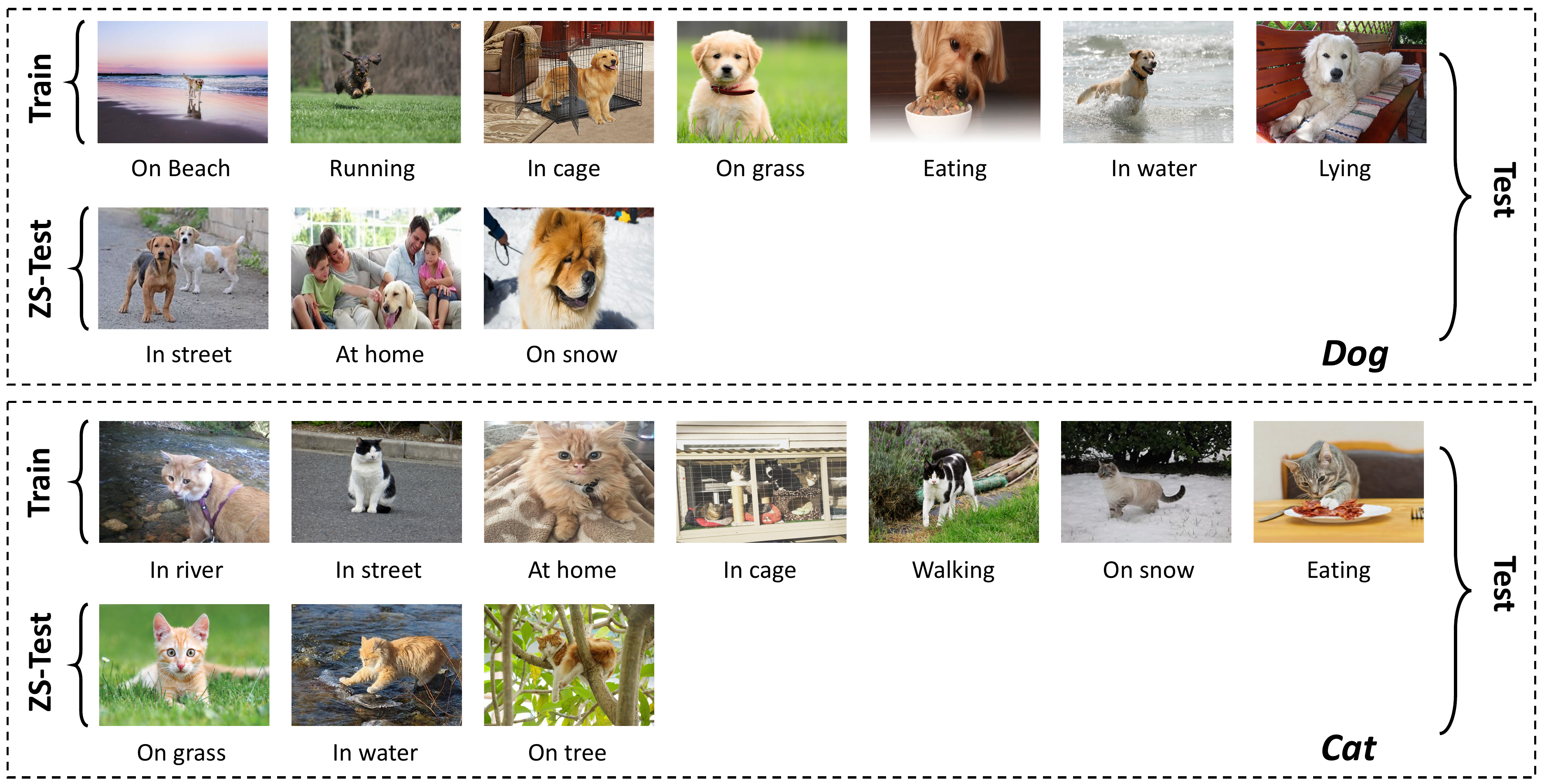}
\end{center}
  \caption{We list the sample images of each context class using ``Dog'' and ``Cat'' as the example in our constructed NICO dataset. \textbf{Train}, \textbf{Test} and \textbf{ZS-Test} denote samples for training, testing and zero shot testing respectively. Note that there is no overlap between training and testing images.}
\label{fig:nico_dataset}
\end{figure*}

%% file: appendix/figures/visualize_group_full_balance.tex
\begin{figure}[t!]
    \captionsetup{font=footnotesize,labelfont=footnotesize}
    \centering
    \includegraphics[width=1.0\linewidth]{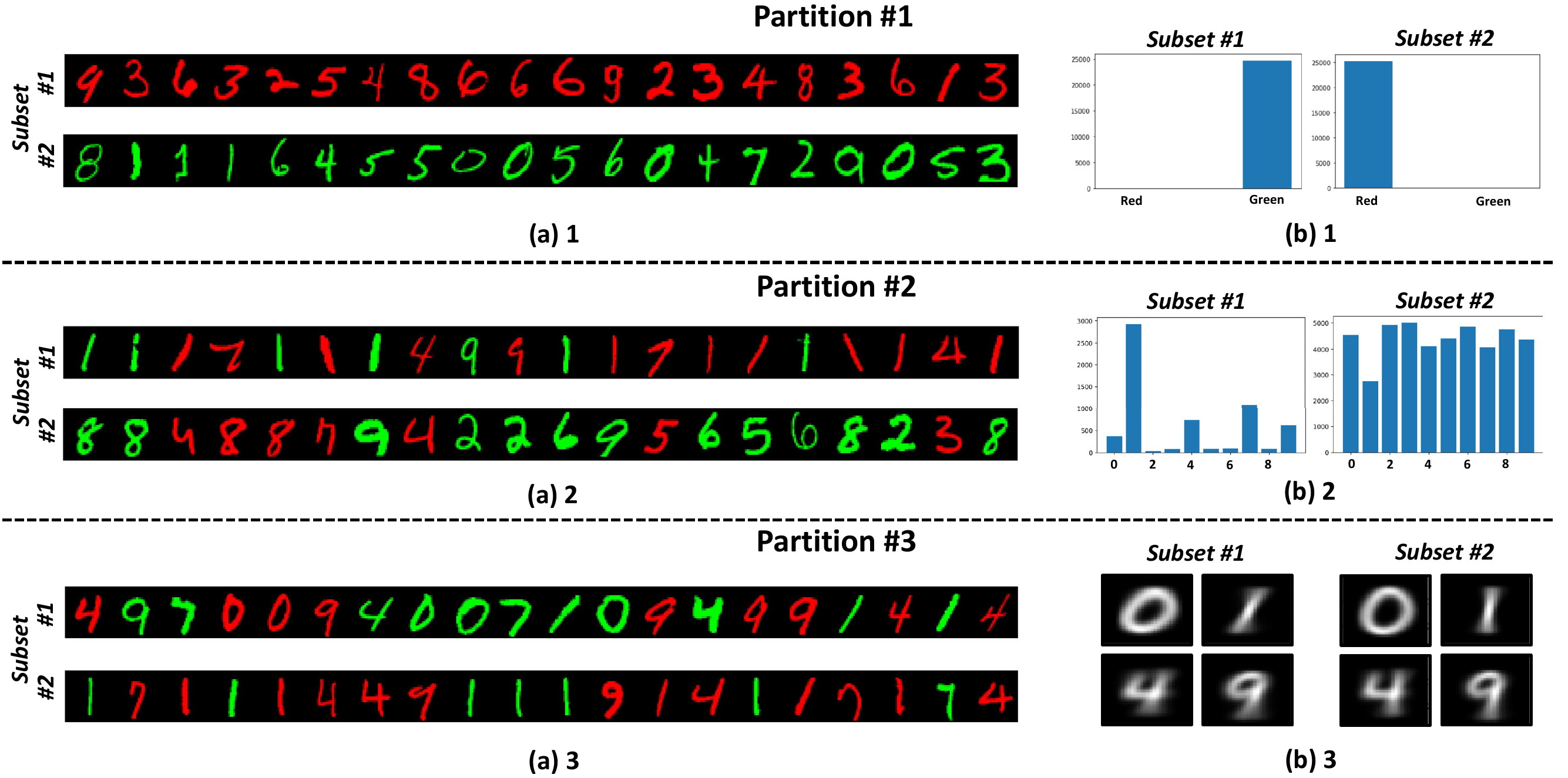}
    \caption{Visualization of the obtained partition $\mathbf{P^{*}}$ during training on the full CMNIST dataset with balanced colorization. The left part (\ie, (a)1, (a)2, (a)3) is the 20 samples chosen randomly from each specific subset in different partitions, similar to the Figure~5 in the main paper. Note that for (a)3 we mainly show 0,1,4,7,9 for ease of comparing.
    The right part ((b)1, (b)2, (b)3) is the global statistic view of the two subsets. Specifically, (b)1 plots the number of images with different colors in subset \#1 and \#2; (b)2 draws the number of images in terms of the digit in subset \#1 and \#2; (b)3 shows the images by averaging samples from the same digit.}
    \label{fig:visual_group_full_balance}
    \vspace{-2mm}
\end{figure}

%% file: appendix/tables/ssl_imagenet100.tex
\begin{wraptable}{r}{0.6\textwidth}
\centering
\captionsetup{font=footnotesize,labelfont=footnotesize,skip=5pt}
\setlength\extrarowheight{1pt}
\scalebox{0.85}{
\begin{tabular}{m{3cm} p{0.9cm}<{\centering}p{0.9cm}<{\centering}p{0.9cm}<{\centering}p{0.9cm}<{\centering}}
\hline\hline
\multicolumn{1}{c}{\multirow{3}[2]{*}{\large{Method}}} &  \multicolumn{4}{c}{\textbf{ImageNet100}}  \\ 
& \multicolumn{2}{c}{\textbf{$k$-NN}} & \multicolumn{2}{c}{\textbf{Linear}} \\
\cmidrule(lr){2-3}\cmidrule(lr){4-5}
& \textit{Top-1} & \textit{Top-5} & \textit{Top-1} & \textit{Top-5}\\ \hline

    SimCLR~\cite{chen2020simple} & 64.64 & 89.36 & 76.58 & 94.22 \\
    
    DCL~\cite{chuang2020debiased} & 65.12 & 89.86 & 77.32 & 94.74 \\
    
    HCL~\cite{kalantidis2020hard} & 69.24 & 90.66 & 79.15 & 94.72 \\

    \hline

    \textbf{SimCLR+IP-IRM} & \cellcolor{mygray} 69.08 & \cellcolor{mygray} 90.96 & \cellcolor{mygray} 79.52 & \cellcolor{mygray} 94.76 \\
    
    \textbf{DCL+IP-IRM} & \cellcolor{mygray} 69.62 & \cellcolor{mygray} 91.62 & \cellcolor{mygray} \textbf{80.38} & \cellcolor{mygray} 95.32 \\
    
    \textbf{HCL+IP-IRM} & \cellcolor{mygray} \textbf{71.78} & \cellcolor{mygray} \textbf{91.94} & \cellcolor{mygray} 80.30 & \cellcolor{mygray} \textbf{95.38} \\

\hline \hline
\end{tabular}}
\caption{Accuracy (\%) of k-NN and linear classifiers trained on the representations learnt on ImageNet100~\cite{tian2019contrastive} with 200 training epochs. We used the pretext task in SimCLR~\cite{chen2020simple}, DCL~\cite{chuang2020debiased} and HCL~\cite{kalantidis2020hard} with IP-IRM, denoted as $(\cdot) + \text{IP-IRM}$.}
\label{tab:appx_ssl_small}
\vspace{-10mm}
\end{wraptable}

%% file: appendix/appx_sections/exp.tex
\section{Additional Experimental Results}
\label{sec:exp}

\subsection{Visualization of Partition $\mathbf{P^{*}}$}

As we proposed in Section 5.1, here we plot the partition results on the full CMNIST dataset with balanced colorization (each image is uniformly colored by red or green) in Figure~\ref{fig:visual_group_full_balance}. We can see in each partition, the obtained $\mathbf{P^*}$ still tells apart a specific semantic into two subsets. Similar to that in the experiment on CMNIST binary dataset (see Figure~5 (a) in the main paper), the semantics of color, digit and slant can be obviously discovered by our IP-IRM algorithm.

\subsection{Unsupervised Disentanglement on Full Shapes3D Dataset}

As we introduced in Section 5.1, we evaluate our IP-IRM and baseline models on Shapes3D dataset with only first three semantics, as the standard augmentations in SSL will contaminate any color-related semantics. Here we present the results on full Shapes3D dataset in Table~\ref{tab:disent_full}. We can find that the VAE-based models indeed perform much better than the SSL model due to the well-dientangled color semantics.

\input{appendix/tables/disentangle_full}

\subsection{Additional Results on ImageNet100}

We additionally performed our algorithm and baselines on the medium-scale dataset --- ImageNet100. Similar to the results presented in the Table~2 of the main paper, incorporating IP-IRM algorithm to the baselines still brings the huge performance boosts to both $k$-NN and linear classification. 
For example, we can observe that our IP-IRM improves the SimCLR by 4.44\% on $k$-NN Top-1 and 2.94\% on linear classification Top-1 accuracy. Moreover, our DCL+IP-IRM achieves a new state-of-the-art accuracy of 80.38\% in the downstream linear classification evaluation.
Also we can find that IP-IRM brings more performance gain with $k$-NN classifier compared to the supervised linear classifier, \eg, 4.50\% using DCL+IP-IRM and 2.54\% with HCL+IP-IRM. This validates that the constructed feature space using IP-IRM more faithfully reflect the semantic differences than the baselines.

\subsection{Detailed Results of Transfer Learning}

\input{appendix/tables/transfer}

As we wrote in the main paper that we also conducted experiments of many-shot and 20-shot, here we show the results in Table~\ref{tab:appx_transfer}. 
From the results we can make the following observations: (i) The downstream transfer performance is approximately correlated to the in-domain self-supervised learning evaluation, \ie, the linear classification accuracy. The better methods in ImageNet classification can also obtain higher transfer performance.
(ii) Our IP-IRM outperforms all the counterparts on most datasets and achieves best performance on the Avg..
(iii) Combining with the 5-shot results in Table~4 of the main paper, we can find with the increasing training samples (\ie, 5-shot $\rightarrow$ 20-shot $\rightarrow$ many-shot), the performance gain of our IP-IRM decreases. This is in line with the conclusion of~\cite{van2019disentangled} that the disentangled feature is more helpful for the learning with fewer training samples.

\input{appendix/tables/transfer_std}

To report the error bars as stated in the checklist, here we presented the standard deviation for the 5-shot experiments as a supplement for the Table~4 in the main paper. As shown in Table~\ref{tab:transfer_std}, all the methods perform in a stable interval.

\subsection{Interventional Few-Shot Learning}
We evaluated the learned representations by SSL and our IP-IRM with the IFSL classifier proposed in~\cite{yue2020interventional}, which is an intervention-based classifier designed to remove the confounding bias from the misuse of pre-trained knowledge in few-shot learning (\eg, treating grass as dog when most dogs appear with grass in the few-shot training images). We adopted the combined adjustment implementation with $n=4$ (dividing feature channels into 4 subsets). For all models in all datasets, we trained the IFSL classifier using the Adam optimizer with learning rate as 0.01 for 50 epochs. The batch size was set as 4. The reported accuracy is based on the average across 2,000 episodes.

\input{appendix/tables/ifsl}

From the results shown in Table~\ref{tab:ifsl}, we can observe that: i) our IP-IRM obtains the best accuracy, validating the superiority of the learned disentangle representation. ii) IFSL significantly improves the performance of the vanilla linear classifier, showing that IFSL can further deconfound.
iii) Compared with linear classifier, the performance gain of IP-IRM over the best baseline (MoCo-v2) is not as high (1.19\% v.s. 2.22\%). This is reasonable, since the disentangled feature with our proposed IP-IRM helps the downstream classifier to deconfound by neglecting the non-discriminative features (see Section 5.3).

%% file: appendix/tables/disentangle_full.tex
\begin{table}[t!]
\centering
\captionsetup{font=footnotesize,labelfont=footnotesize,skip=5pt}
\setlength\extrarowheight{1pt}
\scalebox{0.75}{
\begin{tabular}{p{0.5cm}<{\centering} p{2.7cm} p{1.6cm}<{\centering}p{1.6cm}<{\centering}p{1.6cm}<{\centering}p{1.6cm}<{\centering}p{1.6cm}<{\centering}p{1.6cm}<{\centering}p{1.6cm}<{\centering}}
\hline\hline
& Method & DCI & IRS & MOD & EXP & LR & GBT & Average\\
\hline

\multicolumn{1}{c}{\multirow{7}{*}{\rotatebox{90}{\textbf{Shapes3D (Full)}}}} & VAE~\cite{kingma2014ICLR} & 0.473\scalebox{.8}{$\pm$0.021} & 0.525\scalebox{.8}{$\pm$0.009} & \textbf{0.907}\scalebox{.8}{$\pm$0.009} & 0.895\scalebox{.8}{$\pm$0.025} & \textbf{0.753}\scalebox{.8}{$\pm$0.072} & 0.472\scalebox{.8}{$\pm$0.021} & 0.671\scalebox{.8}{$\pm$0.015}\\
& $\beta$-VAE~\cite{higgins2016beta} & 0.495\scalebox{.8}{$\pm$0.023} & 0.485\scalebox{.8}{$\pm$0.032} & 0.859\scalebox{.8}{$\pm$0.026} & \textbf{0.899}\scalebox{.8}{$\pm$0.006} & 0.713\scalebox{.8}{$\pm$0.013} & 0.496\scalebox{.8}{$\pm$0.025} & 0.658\scalebox{.8}{$\pm$0.012}\\
& $\beta$-AnnealVAE~\cite{burgess2018understanding} & \textbf{0.634}\scalebox{.8}{$\pm$0.033} & \textbf{0.813}\scalebox{.8}{$\pm$0.050} & 0.758\scalebox{.8}{$\pm$0.095} & 0.786\scalebox{.8}{$\pm$0.024} & 0.598\scalebox{.8}{$\pm$0.034} & \textbf{0.634}\scalebox{.8}{$\pm$0.033} & \textbf{0.704}\scalebox{.8}{$\pm$0.025}\\
& $\beta$-TCVAE~\cite{chen2018isolating} & 0.556\scalebox{.8}{$\pm$0.048} & 0.524\scalebox{.8}{$\pm$0.074} & 0.797\scalebox{.8}{$\pm$0.054} & 0.897\scalebox{.8}{$\pm$0.028} & 0.718\scalebox{.8}{$\pm$0.042} & 0.569\scalebox{.8}{$\pm$0.072} & 0.677\scalebox{.8}{$\pm$0.048}\\
& Factor-VAE~\cite{kim2018disentangling} & 0.454\scalebox{.8}{$\pm$0.013} & 0.447\scalebox{.8}{$\pm$0.027} & 0.798\scalebox{.8}{$\pm$0.032} & 0.839\scalebox{.8}{$\pm$0.030} & 0.637\scalebox{.8}{$\pm$0.029} & 0.455\scalebox{.8}{$\pm$0.013} & 0.605\scalebox{.8}{$\pm$0.007} \\ \cmidrule(lr){2-9}
& SimCLR~\cite{chen2020simple} & 0.368\scalebox{.8}{$\pm$0.020} & 0.452\scalebox{.8}{$\pm$0.031} & 0.795\scalebox{.8}{$\pm$0.034} & 0.796\scalebox{.8}{$\pm$0.008} & 0.516\scalebox{.8}{$\pm$0.055} & 0.369\scalebox{.8}{$\pm$0.019} & 0.549\scalebox{.8}{$\pm$0.027}\\
& \textbf{IP-IRM (Ours)} & \cellcolor{mygray}0.392\scalebox{.8}{$\pm$0.010} & \cellcolor{mygray}0.426\scalebox{.8}{$\pm$0.011} & \cellcolor{mygray} 0.835\scalebox{.8}{$\pm$0.024} & \cellcolor{mygray} 0.806\scalebox{.8}{$\pm$0.006} & \cellcolor{mygray} 0.523\scalebox{.8}{$\pm$0.018} & \cellcolor{mygray} 0.391\scalebox{.8}{$\pm$0.012} & \cellcolor{mygray} 0.562\scalebox{.8}{$\pm$0.012}\\

\hline \hline
\end{tabular}}
\caption{Results on disentanglement metrics of existing unsupervised disentanglement methods, standard SSL (SimCLR~\cite{chen2020simple}) and IP-IRM using Shapes3D~\cite{kim2018disentangling} (full). Results are averaged over 4 trails (mean $\pm$ std).}
\label{tab:disent_full}
\vspace*{-6mm}
\end{table}

%% file: appendix/tables/transfer.tex
\begin{table}[h!]
\centering
\captionsetup{font=footnotesize,labelfont=footnotesize,skip=5pt}
\setlength\extrarowheight{1pt}
\setlength\tabcolsep{4pt}
\scalebox{0.8}{
\begin{tabular}{p{0.1cm}<{\centering} p{2.5cm} p{1.0cm}<{\centering}p{1.0cm}<{\centering}p{0.8cm}<{\centering}p{1.0cm}<{\centering}p{1.1cm}<{\centering}p{0.8cm}<{\centering}p{1.0cm}<{\centering}p{0.8cm}<{\centering}p{0.8cm}<{\centering}p{0.8cm}<{\centering}p{0.8cm}<{\centering}p{0.7cm}<{\centering}}
\hline\hline
\multicolumn{2}{c}{Method} & Aircraft & Caltech & Cars & Cifar10 & Cifar100 & DTD & Flowers & Food & Pets & SUN & VOC & Avg.\\
\hline

\multicolumn{1}{c}{\multirow{7}{*}{\rotatebox{90}{\textbf{Many-Shot}}}} & InsDis~\cite{wu2018unsupervised} & 36.87 & 71.12 & 28.98 & 80.28 & 59.97 & 68.46 & 83.44 & 63.39 & 68.78 & 49.47 & 74.37 & 62.29\\
& PCL~\cite{li2020prototypical} & 21.61 & 76.90 & 12.93 & 81.84 & 55.74 & 62.87 & 64.73 & 48.02 & 75.34 & 45.70 & 78.31 & 56.73\\
& PIRL~\cite{misra2020self} & 37.08 & 74.48 & 28.72 & 82.53 & 61.26 & 68.99 & 83.60 & 64.65 & 71.36 & 53.89 & 76.61 & 63.92\\
& MoCo-v1~\cite{he2019moco} & 35.55 & 75.33 & 27.99 & 80.16 & 57.71 & 68.83 & 82.10 & 62.10 & 69.84 & 51.02 & 75.93 & 62.41\\
& MoCo-v2~\cite{chen2020mocov2} & 41.28 & \textbf{87.91} & 40.04 & 91.33 & 73.14 & \textbf{74.47} & 89.02 & 67.10 & 80.49 & 58.10 & 80.13 & 71.18\\

& \textbf{IP-IRM (Ours)} & \cellcolor{mygray}\textbf{43.12} & \cellcolor{mygray} 87.22 & \cellcolor{mygray} \textbf{41.16} & \cellcolor{mygray} \textbf{91.84} & \cellcolor{mygray} \textbf{74.13} & \cellcolor{mygray} 73.94 & \cellcolor{mygray} \textbf{89.23} & \cellcolor{mygray} \textbf{68.05} & \cellcolor{mygray} \textbf{81.70} & \cellcolor{mygray} \textbf{58.41} & \cellcolor{mygray} \textbf{80.32} & \cellcolor{mygray} \textbf{71.74}\\
\hline

\multicolumn{1}{c}{\multirow{7}{*}{\rotatebox{90}{\textbf{20-Shot}}}} 
& InsDis~\cite{wu2018unsupervised} & 47.44 & 88.32 & 54.37 & 67.60 & 72.79 & 82.37 & 92.98 & 69.49 & 82.84 & 90.08 & - & 74.82\\
& PCL~\cite{li2020prototypical} & 44.72 & 92.42 & 47.55 & 69.13 & 70.95 & 79.55 & 81.82 & 67.57 & 92.30 & 90.25 & - & 73.63\\
& PIRL~\cite{misra2020self} & \textbf{48.15} & 90.01 & 55.20 & 67.87 & 73.27 & 82.83 & 92.89 & 69.98 & 84.43 & 90.54 & - & 75.52\\
& MoCo-v1~\cite{he2019moco} & 47.08 & 90.81 & 51.98 & 64.01 & 69.72 & 83.26 & 92.54 & 70.12 & 83.78 & 90.00 & - & 74.33\\
& MoCo-v2~\cite{chen2020mocov2} & 42.57 & 95.75 & 57.30 & 73.99 & 79.20 & \textbf{86.83} & 93.15 & 73.64 & 91.59 & 93.85 & - & 78.78\\
& \textbf{IP-IRM (Ours)} & \cellcolor{mygray} 43.92 & \cellcolor{mygray} \textbf{95.77} & \cellcolor{mygray} \textbf{57.59} & \cellcolor{mygray} \textbf{76.43} & \cellcolor{mygray} \textbf{81.56} & \cellcolor{mygray} 86.69 & \cellcolor{mygray} \textbf{94.03} & \cellcolor{mygray} \textbf{74.81} & \cellcolor{mygray} \textbf{92.35} & \cellcolor{mygray} \textbf{94.19} & \cellcolor{mygray} - & \cellcolor{mygray} \textbf{79.73}\\

\hline \hline
\end{tabular}}
\caption{Accuracy (\%) of transfer learning experiments using representation trained on ImageNet~\cite{deng2009imagenet}. Few-shot experiment was conducted with 5-way-20-shot setting using 2,000 episodes. We excluded VOC~\cite{everingham2010pascal} in few-shot experiments following~\cite{Ericsson2021HowTransfer} as it is a multi-label dataset where standard few-shot evaluation is not applicable.}
\label{tab:appx_transfer}
\vspace{-0.3cm}
\end{table}

%% file: appendix/tables/transfer_std.tex
\begin{table}[h!]
\centering
\captionsetup{font=footnotesize,labelfont=footnotesize,skip=5pt}
\setlength\extrarowheight{1pt}
\setlength\tabcolsep{4pt}
\scalebox{0.8}{
\begin{tabular}{p{2.5cm} p{2.0cm}<{\centering}p{2.0cm}<{\centering}p{2.0cm}<{\centering}p{2.0cm}<{\centering}p{2.1cm}<{\centering}}
\hline\hline
Method & Aircraft & Caltech & Cars & Cifar10 & Cifar100 \\
\hline

InsDis~\cite{wu2018unsupervised} & 35.07$\pm$0.43 & 75.97$\pm$0.47 & 37.49$\pm$0.36 & 51.49$\pm$0.40 & 57.61$\pm$0.48 \\
PCL~\cite{li2020prototypical} & \textbf{36.86}$\pm$0.44 & 90.72$\pm$0.30 & 39.68$\pm$0.39 & 59.26$\pm$0.36 & 60.78$\pm$0.46 \\
PIRL~\cite{misra2020self} & 36.70$\pm$0.44 & 78.63$\pm$0.46 & 39.21$\pm$0.37 & 49.85$\pm$0.42 & 55.23$\pm$0.52 \\
MoCo-v1~\cite{he2019moco} & 35.31$\pm$0.44 & 79.60$\pm$0.45 & 36.35$\pm$0.35 & 46.96$\pm$0.39 & 51.62$\pm$0.50 \\
MoCo-v2~\cite{chen2020mocov2} & 31.98$\pm$0.38 &  92.32$\pm$0.29 & 41.47$\pm$0.41 & 56.50$\pm$0.42 & 63.33 $\pm$0.51\\
\textbf{IP-IRM (Ours)} & \cellcolor{mygray} 32.98$\pm$0.40 & \cellcolor{mygray} \textbf{93.16}$\pm$0.26 & \cellcolor{mygray} \textbf{42.87}$\pm$0.41 & \cellcolor{mygray} \textbf{60.73}$\pm$0.39 & \cellcolor{mygray} \textbf{68.54}$\pm$0.49 \\

\hline \hline
Method & DTD & Flowers & Food & Pets & SUN\\
\hline

InsDis~\cite{wu2018unsupervised}  & 69.38$\pm$0.43 & 77.35$\pm$0.51 & 50.01$\pm$0.45 & 66.38$\pm$0.44 & 74.97$\pm$0.48 \\
PCL~\cite{li2020prototypical} & 69.53$\pm$0.43 & 67.50$\pm$0.50 & 57.06$\pm$0.44 & \textbf{88.31}$\pm$0.36 & 84.51$\pm$0.37 \\
PIRL~\cite{misra2020self} & 70.43$\pm$0.42 & 78.37$\pm$0.48 & 51.61$\pm$0.45 & 69.40$\pm$0.43 & 76.64$\pm$0.46 \\
MoCo-v1~\cite{he2019moco} & 68.76$\pm$0.46 & 75.42$\pm$0.53 & 49.77$\pm$0.46 & 68.32$\pm$0.43 & 74.77$\pm$0.50 \\
MoCo-v2~\cite{chen2020mocov2} & 78.00$\pm$0.38 & 80.05$\pm$0.45 & 57.25$\pm$0.48 & 83.23$\pm$0.40 & 88.10$\pm$0.33  \\
\textbf{IP-IRM (Ours)} & \cellcolor{mygray} \textbf{79.30}$\pm$0.36 & \cellcolor{mygray} \textbf{82.68}$\pm$0.41 & \cellcolor{mygray} \textbf{59.61}$\pm$0.46 & \cellcolor{mygray} 85.23$\pm$0.38 & \cellcolor{mygray} \textbf{89.38}$\pm$0.30  \\
\hline \hline

\end{tabular}}
\caption{Accuracy (\%) of 5-way-5-shot few-shot evaluation with standard deviation (mean$\pm$std) using the image representation learned on ImageNet~\cite{deng2009imagenet}. }
\label{tab:transfer_std}
\vspace{-6mm}
\end{table}

%% file: appendix/tables/ifsl.tex
\begin{table}[h!]
\centering
\captionsetup{font=footnotesize,labelfont=footnotesize,skip=5pt}
\setlength\extrarowheight{1pt}
\setlength\tabcolsep{4pt}
\scalebox{0.8}{
\begin{tabular}{p{0.1cm}<{\centering} p{2.5cm} p{1.0cm}<{\centering}p{1.0cm}<{\centering}p{0.8cm}<{\centering}p{1.0cm}<{\centering}p{1.1cm}<{\centering}p{0.8cm}<{\centering}p{1.0cm}<{\centering}p{0.8cm}<{\centering}p{0.8cm}<{\centering}p{0.8cm}<{\centering}p{0.7cm}<{\centering}}
\hline\hline
\multicolumn{2}{c}{Method} & Aircraft & Caltech & Cars & Cifar10 & Cifar100 & DTD & Flowers & Food & Pets & SUN & Avg.\\
\hline

\multicolumn{1}{c}{\multirow{6}{*}{\rotatebox{90}{\textbf{Linear}}}} & InsDis~\cite{wu2018unsupervised} & 35.07 & 75.97 & 37.49 & 51.49 & 57.61 & 69.38 & 77.35 & 50.01 & 66.38 & 74.97 & 59.57\\
& PCL~\cite{li2020prototypical} & \textbf{36.86} & 90.72 & 39.68 & 59.26 & 60.78 & 69.53 & 67.50 & 57.06 & \textbf{88.31} & 84.51 & 65.42\\
& PIRL~\cite{misra2020self} & 36.70 & 78.63 & 39.21 & 49.85 & 55.23 & 70.43 & 78.37 & 51.61 & 69.40 & 76.64 & 60.61\\
& MoCo-v1~\cite{he2019moco} & 35.31 & 79.60 & 36.35 & 46.96 & 51.62 & 68.76 & 75.42 & 49.77 & 68.32 & 74.77 & 58.69\\
& MoCo-v2~\cite{chen2020mocov2} & 31.98 &  92.32 & 41.47 & 56.50 & 63.33 & 78.00 & 80.05 & 57.25 & 83.23 & 88.10  & 67.22\\
& \textbf{IP-IRM (Ours)} & \cellcolor{mygray} 32.98 & \cellcolor{mygray} \textbf{93.16} & \cellcolor{mygray} \textbf{42.87} & \cellcolor{mygray} \textbf{60.73} & \cellcolor{mygray} \textbf{68.54} & \cellcolor{mygray} \textbf{79.30} & \cellcolor{mygray} \textbf{82.68} & \cellcolor{mygray} \textbf{59.61} & \cellcolor{mygray} 85.23 & \cellcolor{mygray} \textbf{89.38}  & \cellcolor{mygray} \textbf{69.44}\\

\hline

\multicolumn{1}{c}{\multirow{6}{*}{\rotatebox{90}{\textbf{IFSL}}}} & InsDis~\cite{wu2018unsupervised} & 41.49 & 86.53 & 45.37 & 60.51 & 66.94 & 75.99 & 87.61 & 58.85 & 80.12 & 85.87 & 68.93\\
& PCL~\cite{li2020prototypical} & 41.22 & 91.23 & 44.55 & 64.96 & 68.10 & 73.10 & 77.99 & 62.35 & 90.88 & 87.28 & 70.17\\
& PIRL~\cite{misra2020self} & \textbf{42.31} & 88.54 & 46.61 & 61.07 & 66.49 & 76.70 & \textbf{88.20} & 60.37 & 82.59 & 87.01 & 69.99\\
& MoCo-v1~\cite{he2019moco} & 41.43 & 89.30 & 44.20 & 57.01 &63.29  &76.84  &87.12  &60.80  &81.73  &85.90  &68.76 \\
& MoCo-v2~\cite{chen2020mocov2} &40.02  &93.95  &51.12  &69.56  &73.84  &80.67  &86.94  &66.10  &90.38  &91.18  &74.38 \\

& \textbf{IP-IRM (Ours)} & \cellcolor{mygray}41.49 & \cellcolor{mygray} \textbf{94.31} & \cellcolor{mygray} \textbf{52.80} & \cellcolor{mygray} \textbf{72.02} & \cellcolor{mygray} \textbf{76.52} & \cellcolor{mygray} \textbf{80.78} & \cellcolor{mygray} 87.69 & \cellcolor{mygray} \textbf{67.35} & \cellcolor{mygray} \textbf{91.13} & \cellcolor{mygray} \textbf{91.62} & \cellcolor{mygray} \textbf{75.57}\\

\hline \hline
\end{tabular}}
\caption{Accuracy (\%) of IFSL classifier using representation trained on ImageNet~\cite{deng2009imagenet}. Experiment was conducted with 5-way-5-shot setting using 2,000 episodes. We used the combined adjustment for all experiments.}
\label{tab:ifsl}
\vspace{-0.3cm}
\end{table}

%% file: appendix/appx_sections/others.tex
\section{Others}

\subsection{License}

We use the following open-source database and the license just follow them.
\begin{itemize}
    \item \url{https://github.com/YannDubs/disentangling-vae}
    \item \url{https://github.com/chingyaoc/DCL}
    \item \url{https://github.com/joshr17/HCL}
    \item \url{https://github.com/facebookresearch/moco}
    \item \url{https://github.com/taoyang1122/pytorch-SimSiam}
    \item \url{https://github.com/facebookresearch/InvariantRiskMinimization}
    \item \url{https://github.com/kibok90/imix}
\end{itemize}

\subsection{Data Consent}

We entirely use the open-source data and they have already obtained the consent during the data collection. Details can be checked on their website.

\subsection{Personally Identifiable Information}

The open-source data we used does not contain personally identifiable information or offensive context. More information can refer to the website of the open-source data.